\documentclass{article}
\usepackage{arxiv}
\usepackage{booktabs}

\usepackage{lineno}
\usepackage{hyperref}
\usepackage{breakurl}
\modulolinenumbers[5]

\usepackage{graphicx}
\usepackage{amsmath}
\usepackage{multirow}
\usepackage{latexsym}
\usepackage{subcaption}

\usepackage{appendix}
\usepackage{setspace}
\usepackage{ascmac}
\usepackage{natbib}
\bibpunct[:]{[}{]}{,}{a}{}{,}

\title{{\bf \fontsize{20pt}{0pt}\selectfont Dynamic Reinforcement Learning for Actors}}
\author{Katsunari Shibata\\ \\
\texttt{katsunarishibata@gmail.com} \\ \\[-3mm]
Independent Researcher, Kosai, Shizuoka, Japan
}
\begin{document}
\maketitle

\begin{abstract}
%This template helps you to create a properly formatted \LaTeX\ manuscript.
Dynamic Reinforcement Learning (Dynamic RL), proposed in this paper, directly controls system dynamics,
instead of the actor (action-generating neural network) outputs at each moment,
bringing about a major qualitative shift in reinforcement learning (RL) from static to dynamic.
The actor is initially designed to generate chaotic dynamics through the loop with its environment,
enabling the agent to perform flexible and deterministic exploration.
%This allows the agent to explore flexibly without stochastic action selection.

Dynamic RL controls global system dynamics using a local index called ``sensitivity,''
which indicates how much the input neighborhood contracts or expands
into the corresponding output neighborhood through each neuron's processing.
While sensitivity adjustment learning (SAL) prevents excessive convergence of the dynamics,
sensitivity-controlled reinforcement learning (SRL) adjusts them
%—
--- to converge more to improve reproducibility around better state transitions with positive TD error
and to diverge more to enhance exploration around worse transitions with negative TD error.
%This means a leap from general learning, which focuses on moving a point in the state space,
%to an entirely novel type of learning that controls the flow around a line formed by the moving point through time.

Dynamic RL was applied only to the actor in an Actor-Critic RL architecture
while applying it to the critic remains a challenge.
It was tested on two dynamic tasks and
%; a memory-required task and a dynamic pattern generation task.
functioned effectively without external exploration noise or backward computation through time.
Moreover, it exhibited excellent adaptability to new environments, although some problems remain.
%The author believes that it has the potential to be an essential technique for the acquisition of
%higher functions such as thinking in which dynamic learning must be critical.
% after further improvement.

Drawing parallels between `exploration' and `thinking,'
%both need autonomous and multi-step state transitions including unexpected ones
the author hypothesizes that ``exploration grows into thinking through learning''
and believes this RL could be a key technique for the emergence of thinking,
%including inspiration that is beyond common sense but makes sense.
including inspiration that cannot be reconstructed from massive existing text data.
%Finally, he mentions what he thinks about the grave potential risks of this RL despite his presumption.
Finally, despite being presumptuous,
the author presents the argument that this research should not proceed due to its potentially fatal risks,
aiming to encourage discussion.
%he urges researchers not to proceed with the research on Dynamic RL hereafter due to its potentially fatal risks.

$\langle$ Highlights $\rangle$
\begin{itemize}
\item This paper proposes Dynamic Reinforcement Learning (Dynamic RL), which controls system dynamics
generating exploration-embedded motions without stochastic action selection.
\item Dynamic RL comprises two learning methods: Sensitivity Adjustment Learning (SAL)
and Sensitivity-controlled Reinforcement Learning (SRL).
\item SAL prevents excessive convergence by maintaining chaos in the system dynamics.
\item SRL makes the system dynamics more convergent for better reproducibility
or more divergent for more exploration depending on the TD error.
\item Dynamic RL is demonstrated to function as RL without BPTT in two dynamic tasks
and exhibits excellent adaptability to new environments.
\item The author suggests its potential as a fundamental technique for the emergence of thinking
and is terribly concerned about its risks.
\end{itemize}
\end{abstract}

\keywords{
%\texttt{elsarticle.cls}\sep \LaTeX\sep Elsevier \sep template
%\MSC[2010] 00-01\sep  99-00
reinforcement learning (RL), 
%Dynamic Reinforcement Learning (Dynamic RL),
recurrent neural network (RNN),
chaotic dynamics, exploration, sensitivity, thinking
}

%\end{frontmatter}
\pagestyle{fancy}
%\fancyhead{}
%\fancyhead[CE]{Sensitivity\ \  \small{-- Local Index to Control Chaoticity and Gradient Globally --}}
%\fancyhead[CO]{\small{K.~Shibata, T.~Ejima, Y.~Tokumaru, T.~Matsuki}}
\title{Dynamic Reinforcement Learning for Actors \hspace{8.4cm}K.~Shibata}

\begin{itembox}[|]{\bf Statement}
Dynamic Reinforcement Learning (Dynamic RL) proposed in this paper has the potential
to endow Artificial Intelligence (AI) or Artificial General Intelligence (AGI) with the ability to think and discover new things,
which could pose a grave threat to humanity.
Now, while this research is still taking baby steps,
we should halt further progress to protect humanity from the risk.
%for the sake of humanity.
The author sincerely and earnestly urges researchers to refrain from advancing this work
and developers to refrain from developing a library for Dynamic RL in frameworks,
% not to proceed with
at least until a consensus is reached among humans on this matter.
Let us pause, imagine, and contemplate, free from preconceptions before it is too late.
\\ \\
Refer to subsection \ref{subsec:Risk} for the discussion leading to this statement.
\end{itembox}

%\linenumbers

\section{Introduction}\label{Sec:Introduction}
%\section{The Elsevier article class}
Reinforcement learning (RL) has evolved into a core technology for autonomous, non-supervised learning
 in modern artificial intelligence (AI) through several significant qualitative advancements.
As the next major step in its evolution, the author introduces `Dynamic RL.'
Unlike conventional methods, Dynamic RL treats exploration as an inherent aspect of actions
rather than an independent process.
Instead of using external noise for stochastic selection, exploration is generated as chaotic system dynamics
through the action-generating (actor) network, and RL controls these dynamics directly.

\subsection{Evolution of RL toward Learning of All Kinds of Human Functions}
RL initially involved only learning appropriate tables or mappings from a discrete state space
to a discrete action space within a Markov decision process (MDP) \citep{Sutton1998}.
A neural network (NN) was then introduced, providing many degrees of freedom and parameters.
This enabled an agent to learn not only continuous nonlinear mappings but also the entire sensor-to-motor process
based on a value function using reinforcement signals with the help of the gradient-based method.
Consequently, RL has advanced from learning of actions to learning of various functions,
%that are necessary for the sensor-to-motor process,
including recognition and prediction \citep{ICNN97,Intech, End-to-End}.
%What makes RL attractive is the gap that while just a scalar reinforcement signal is given,
%functions that are hard to describe for humans can be acquired autonomously,
%and so it is possible to acquire a variety of functions that humans have not taught.
A notable achievement occurred when an RL agent learned the game strategy of ``tunneling'' in the game ``Breakout''
solely by using changes in the game score as a reinforcement signal \citep{Mnih}.
RL has since become a driving force in achieving human-level, and sometimes superhuman, performance
particularly in gaming \citep{OpenAI-Five, AlphaStar, MuZero}.

Furthermore, the introduction of a recurrent neural network (RNN) has opened the way for RL
to process and learn along the time axis.
This allowed agents to extract necessary information from large amounts of past information
and then retain and utilize it.
This advancement allowed learning in partially observable situations,
such as partially observable MDPs (POMDPs), and solving memory-dependent tasks.
As a result, the functions acquired through RL span a wide range,
including recognition, memory, attention, prediction, and communication \citep{ISADS99,Intech,End-to-End,Communication}.

Does this suggest that RL agents can acquire all functions that humans have, even simple ones, through learning?
The current answer is `No'.
Unfortunately, the author has yet to accept any of these as functions that could be called `thinking.' 
Therefore, he has set the emergence of such a function through learning as his major future target
%also
and has considered what would be required for the emergence for more than a decade.
%The author has yet to encounter the emergence of any function that could be called `thinking'.
%Therefore, he has set the emergence of such a function through learning as his major future target
%and has considered what would be required for the emergence.

\subsection{`Thinking' and Recent Advancements in Generative AI}
Humans are intelligent, and it would not be an exaggeration to say that `thinking' is the most typical form of intelligence.
`Thinking' has been a central topic in philosophy for a long time since the time of Plato \citep{Plato},
with numerous ideas explored \citep{Britannica}.
About 80 years ago, attempts began to explain human intelligence from a computational perspective
by considering the mind or neural networks, which are thought to be the source of intelligence in living organisms,
as computational machines \citep{McCulloch_Pitts,Turing,Newell}.

Recently, generative AI, especially large-scale language models (LLMs) such as GPT \citep{GPT} and Gemini \citep{Gemini},
appear to understand what we say, think about it, and come up with an appropriate response.
We, at least the author, feel(s) as if we are interacting with a human who has intelligence.
It was reported that ChatGPT powered by GPT-3.5 performed near or over the passing threshold
on the United States Medical Licensing Exam (USMLE) \citep{Chat-GPT},
and also GPT-4 passed a simulated bar exam with a score around the top 10\% of test takers \citep{GPT-4}.

Transformer \citep{Transformer}, a central technique of many LLMs, encodes input patterns into a latent space
constructed through learning, capturing complex dependencies primarily relying on self-attention mechanisms,
without relying on sequence-aligned RNNs or convolutional networks.
Then, it generates likely responses in LLMs based on learning from a vast amount of data.
It seems different for the author from the common image of thinking: ``letting one's mind wander.''
Actually, in response to the question ``Are you thinking?'', both Chat-GPT and Gemini themselves deny
that they think like a human.
%"No, I don't ``think'' in the way humans do."
%My responses are generated by analyzing patterns in the input, matching them with my training data,
%and applying algorithms to produce relevant answers. It might seem like I'm thinking,
%but it's purely computational processing, not consciousness or independent thought."
The remarkable abilities of large-scale LLMs, enabled by their excellent scalability,
might suggest that overwhelming quantity or scale changes the quality drastically.
However, the author remains amazed that such a relatively simple underlying technology
is capable of generating such human-like responses.

From the perspective of `thinking,' there are several studies \citep{Koivisto, Guzik} that evaluate LLMs
as comparable to humans even in tasks designed to observe the ability of divergent thinking \citep{Guilford},
such as the AUT (Alternate Uses Task) \citep{AUT, AUT-net} or TTCT (Torrance Test of Creative Thinking) \citep{TTCT}.
However, one study \citep{Aggarwal} reported that AI machines are not identical to humans in terms of the quality of intelligence or thought
but have human-like logical reasoning systems, which are achieved simply through learning and mimicking human abilities.
Another review on creativity in AI \citep{Ismayilzada} also reported,
being supported by many studies highlighting the inferiority of LLMs in lateral thinking \citep{deBono},
originality, abstract reasoning, and related areas \citep{Huang, Jiang, Zhao, Gendron, Mitchell, Chakrabarty, Lu, Tam}, as follows:
The latest AI models are largely capable of producing linguistically and artistically creative outputs
%such as poems, images, and musical pieces,
but struggle with tasks that require creative problem-solving, abstract thinking, and compositionality.
They exhibit strong interpolation and moderate extrapolation capabilities
but are still far from truly inventing a completely new type of creative artifact.

Lateral thinking requires deviating from established common sense and cannot be achieved
as an interpolation or extrapolation.
In \citep{Lu}, it was observed that machine-generated texts contain significantly more semantic and verbatim matches
with existing web texts compared to high-quality human writings.
%the possibility was considered that the LLM's remarkable creativity may have largely originated
%from the creativity of human-written texts on the web.
%They quantified the linguistic creativity of a given text by estimating
%how much of that text can be reconstructed by mixing and matching a vast amount of existing text snippets on the web.
This observation made it more plausible that the LLM's creativity may largely originate
from the creativity of human-written texts on the web.
%the creativity index of professional human writers is significantly higher than that of LLMs.
To improve creativity, methods such as active divergence and creative decoding
were suggested \citep{Ismayilzada, Broad, Franceschelli}.
However, since they are used in the framework of the current generative deep learning,
it would be difficult to avoid this essential problem.

Independent of the current rapid progress in AI,
the author has arrived at a novel technique for RL that is distinct from Transformer-based generative AI.
In his view, this technique has the potential to become an essential technique for the emergence of `thinking'.
%While Transformer-based generative AI continues to improve at a tremendous rate,
%so its shortcomings may also be addressed in the ongoing progress.
Here, putting the Transformer aside for a moment,
the author introduces Dynamic Reinforcement Learning (Dynamic RL), which brings about a major shift in RL
and explains the potential that this shift may fill the gap in thinking between humans and the current generative AI.

\subsection{Exploration and Thinking}
While the exact definition of `thinking' is difficult to pin down, let us explore the more dynamic form of thinking
that we typically associate with the term.
Humans can think even when remaining still with eyes closed and ears covered,
which seems different from other functions like recognition or prediction.
In order to survive as living organisms, we must at least avoid converging and becoming stagnant.
%We must gain diverse experiences for the future, learn from the experiences, and
%think about a variety of things, processing rationally based on the knowledge acquired through learning.
%Furthermore, we sometimes inspire and make discoveries.
Therefore, many would likely agree that autonomous, rational state transitions,
even without external triggers, are required for `thinking.'
However, acquiring multistep state transitions through learning from scratch in RNNs is very difficult \citep{Sawatsubashi}.
In particular, forming autonomous state transitions without external triggers is challenging.
%Furthermore, the recurrence of the computation makes the outputs of the neurons to be their feedback inputs.
%Therefore, learning often changes the feedback inputs largely,
%and that makes the learning more difficult.
Moreover, considering human abilities such as inspiration and discovery, these autonomous state transitions
should not only be rational but also sometimes unexpected.
%The author believes that it is an essential problem that cannot be avoided
%in the AI using the modern RL with an RNN.

\begin{figure}[t]
\centerline{\includegraphics[scale=0.28]{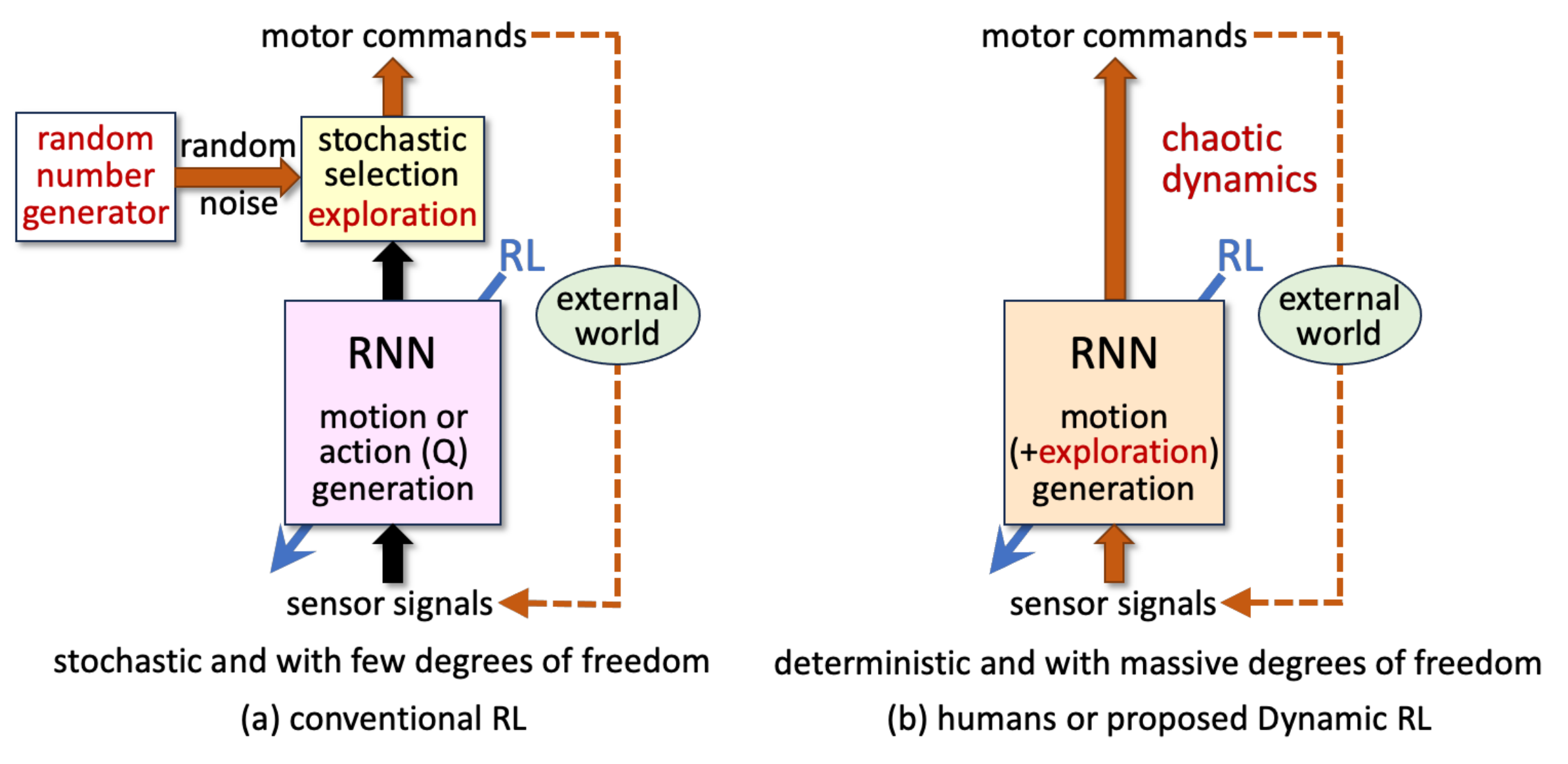}}
\caption{The difference in exploration between conventional RL and humans,
who have inspired the exploration in the proposed Dynamic RL.
In humans or Dynamic RL, the actor RNN embeds exploration factors into motor commands
by inducing chaotic system dynamics without stochastic selection using a random number generator.
(RNN: recurrent neural network)}
\label{fig:Exploration}
\end{figure}

Comparing modern RL with human learning reveals one major difference: `exploration.'
In conventional RL, stochastic selection uses external random noise,
independent of the motion or action generation process, as shown in Fig.~\ref{fig:Exploration}(a).
\footnote{In this paper, the term `motion' is used as continuous, and `action' is used as discrete,
while the term `actor' is used as a generator in both cases.
Action is usually abstract such that the action "turn left" itself is not a motor command,
but is broken down into a series of motor commands for several motors.
Therefore, in the proposed RL, the actor outputs represent continuous motor commands
based on the end-to-end learning concept \citep{End-to-End}.}
For example, $\epsilon$-greedy or Boltzmann selection is applied after computing the Q-value for each action,
neither influencing nor being influenced directly by its computation process.
The exploration noise is not a function of state
even though the temperature is adjusted by simulated annealing \citep{SA}.
%Our human exploration at least does not seem to be stochastic selections only at the final output level,
%which would be the motor level in the sensor-to-motor learning framework.
%Then the author showed that appropriate exploration behaviors themselves
%can be acquired through learning at first \citep{ExplorationLearning1, ExplorationLearning2}.
%However, the agent still needed random exploration to learn the exploration.

In contrast, humans often act while wondering about this and that.
Exploration seems to work during the action or motion generation process.
When we wonder which path to take at a fork in the road, we do not move our hands erratically for exploration.
Exploration is not always uniform; it should depend on the direction in the state space and also the current state.
Furthermore, it should be improved through learning \citep{Expl,Goto_Expl}.
Therefore, the author considers that exploration represents the degree of irregularity in motion generation
and is updated together by the learning of motion generation, as shown in Fig.~\ref{fig:Exploration}(b).
Exploration requires non-convergent, irregular, and unexpected state transitions.
It can be embedded in motor commands by making the system dynamics chaotic.
In other words, by making the dynamics sensitive to minute differences, 
agent behaviors vary even from similar states, leading to the generation of deterministic exploratory state transitions.
%Chaotic dynamics are generated easily by setting the recurrent connection weights in the RNN to be random and large.
Dynamic RL proposed here employs this type of exploration.

\begin{figure}[t]
\centerline{\includegraphics[scale=0.34]{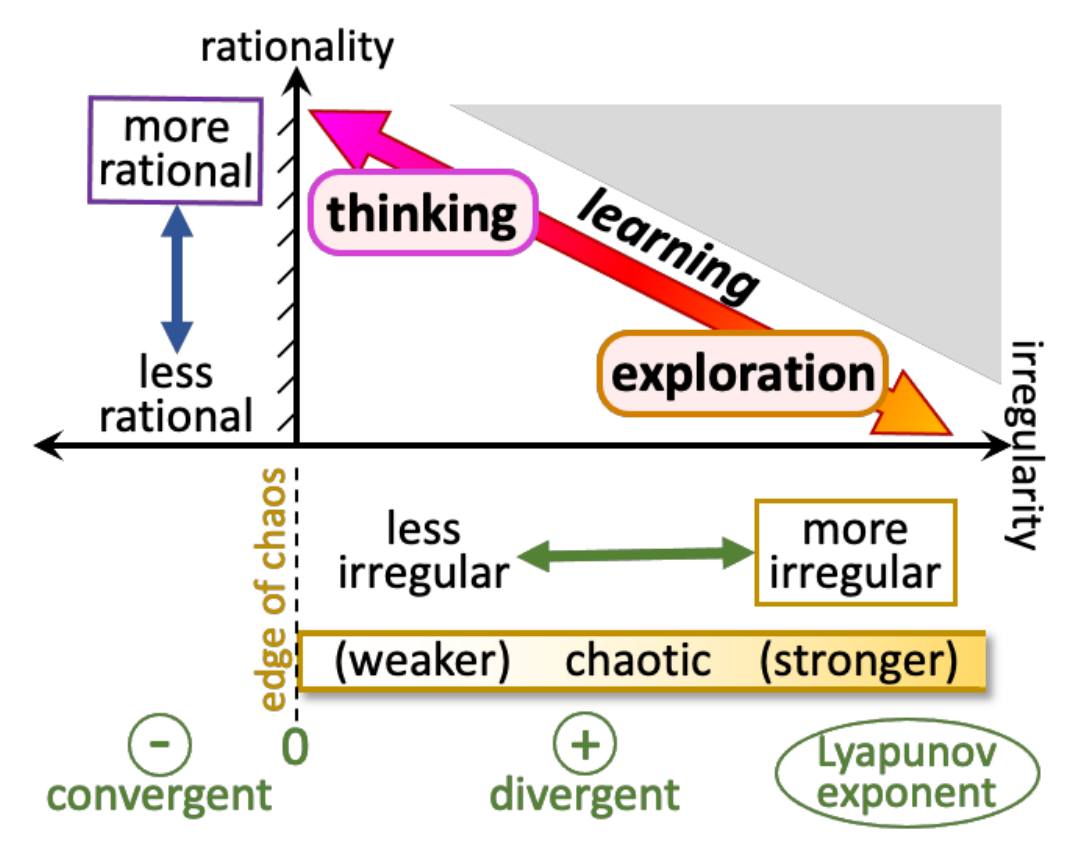}}
\caption{The author's concept of the relationship between `exploration' and `thinking'
 and how they relate to system dynamics.
`Thinking' and `exploration' are similar and inseparable in that both require multistep autonomous state transitions,
and they exist continuously on a spectrum characterized by chaotic dynamics.
In `exploration,' the state transitions must be irregular.
In `thinking,' the state transitions must not only be less irregular but also rational.
Dynamic RL controls the system dynamics based on a value function using reinforcement signals
while preserving chaotic dynamics.}
\label{fig:Thinking}
\end{figure}

%On the other hand, thinking needs autonomous state transitions and also
%unexpected but rational transitions.
%Such transitions also do not emerge with convergent nor strong chaotic system dynamics
%but emerge with weak chaotic dynamics.
%When we see the learning agent as a creature, it cannot explore with convergent dynamics
%and is therefore left with no choice but to die.
From the above, `exploration' is similar to `thinking' in terms of autonomous state transitions,
including unexpectedness.
Both require chaotic system dynamics with a positive Lyapunov exponent rather than convergent dynamics.
The aforementioned points suggest that `exploration' and `thinking' cannot be clearly separated
and may continuously exist along a line within the range of chaotic dynamics.
%Exploration, as well as thinking, does not emerge with converging dynamics
%but requires autonomous state transitions.
In `exploration,' state transitions need to be more irregular with stronger chaotic dynamics.
In contrast, those in `thinking' need to be less irregular but still non-convergent, with weaker chaotic dynamics;
furthermore, they must be more rational.
This means that as shown in Fig.~\ref{fig:Thinking},
they exist on a diagonal line in the two-dimensional space with irregularity and rationality as axes.
If the state transitions are very irregular, they cannot be rational.
In other words, they cannot exist in the upper right portion of Fig.~\ref{fig:Thinking}.
Therefore, ``being less irregular'' is a necessary condition for ``being more rational.''

If the system dynamics are highly chaotic, the agent can explore widely.
The agent cannot `think' before acquiring basic knowledge, such as various cause-and-effect relationships in the world.
Converging the flows around better state transitions makes the transitions more rational,
thereby, the agent acquires the necessary basic knowledge.
That is expected to Dynamic RL.
If autonomous state transitions have already been formed as explorations,
just adjusting the convergence or divergence of transitions is enough to achieve autonomous and rational transitions.
Thus, this should be much easier than creating such transitions from scratch by RL using an RNN with convergent dynamics.
Then, the author posits the following bold hypothesis: ``\textbf{Exploration grows into thinking through learning}''
by shifting state transitions from ``more irregular'' to ``more rational'' while maintaining system dynamics chaotic.
%In thinking, the dynamics of the RNN still need to be chaotic.
Furthermore, when situations change, and so expected rewards cannot be obtained,
Dynamic RL is expected to resume more exploratory behaviors autonomously in the agent.

%Next, let us consider learning methods more concretely.
%In conventional RL, when continuous motions are learned, they are selected stochastically and
%trained based on the product of the difference between actual and expected changes in the value function
%and the exploration noise, which is the difference between actual and default motions.
%However??????????, stochastic exploration is an assumption of this learning method.
\subsection{Learning of Local Dynamics}
In conventional RL, when continuous motions are learned, if the TD error is positive, learning is performed
to move the actor outputs in the direction of the actual motion chosen from the probabilistic distribution.
Conversely, if the TD error is negative, learning tries to move the actor outputs in the opposite direction.
The weights and biases in the network are trained using a gradient-based method.
The author has been uncomfortable with the fact that even though the output computation process of RNNs is dynamic,
learning is still stuck in a static form.
%Moreover, the author has been uncomfortable with the fact that, although RNNs perform dynamic processing,
%their learning methods are still based on static approaches like backpropagation through time (BPTT).
He has tried a major shift in learning from static to dynamic.
This will be discussed in more detail in the subsection \ref{subsec:Future}.
The author's group previously proposed a new method utilizing chaotic dynamics
without assuming stochastic exploration \citep{IJCNN2015},
but the learning of connection weights between hidden neurons did not improve learning performance.

As mentioned, current LLMs are reported to struggle with lateral thinking,
which requires stepping beyond common sense.
%One possible reason for this limitation is that LLMs operate like   or extrapolation
%of the vast training data \citep{Ismayilzada}.
Expecting novelty is analogous to the exploration in RL.
In LLMs, stochastic selections are used,
and the temperature parameter is often expected as a control variable for randomness,
with higher temperatures promoting randomness, and is expected as the creative parameter \citep{Peeperkorn}.
Nevertheless, it has been reported that the temperature does not always work as expected \citep{Peeperkorn,Ismayilzada}.

%Regarding the issues with this expectation, the author offers the following perspective.
It is clear that if state transitions are merely random or irregular, the process can be `exploration' but cannot be `thinking.'
%it does not evoke a sense of novelty.
Even when perceiving novelty in thinking, there must also be some rationality.
However, it seems quite natural that as irregularity increases, rationality decreases.
That is also suggested by the diagonal block arrow in Fig.~\ref{fig:Thinking}.
Actually, a trade-off between novelty and coherence was observed in \citep{Peeperkorn}.
In \citep{Nath}, it was pointed out that LLMs were biased towards either persistence (deep search)
or flexibility (broad search), regardless of parameter settings such as temperature,
while in humans, both persistent and flexible pathways were observed.
Then, how can humans solve this problem even though it seems impossible to solve at first glance?

%%%and they become more unexpected and irregular by making the dynamics more divergent or chaotic in general.

\begin{figure}[t]
\centerline{\includegraphics[scale=0.25]{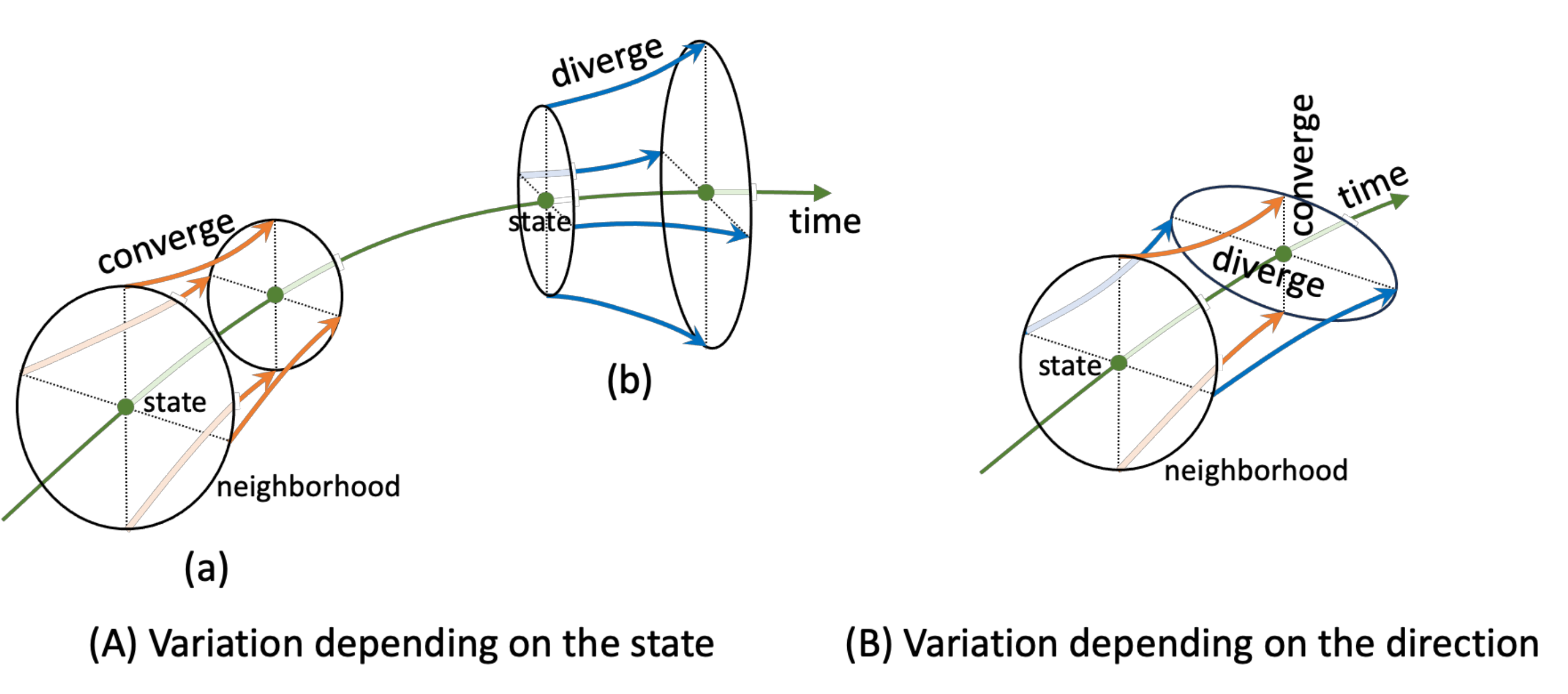}}
\caption{A conceptual diagram explaining the degrees of freedom (DOFs) that dynamics have for a sample case of three-dimensional state space as an example.
(A) Convergence (a) or divergence (b) may vary depending on the state even in the same system.
(B) Convergence or divergence can be varied depending on the direction even in the same state.
For easy viewing, the neighborhood around a state has originally three dimensions, but only two dimensions are presented.
Additionally, the directions of the two eigenvectors are assumed to be orthogonal to each other and also to the direction of state transition.}
\label{fig:LocalDynamics}
\end{figure}

%Following the idea in Fig.~\ref{fig:Thinking}, to achieve `thinking,' the dynamics should be ``more rational'' and ``less irregular''
%but ``chaotic.''
What we should be aware of here is the difference in whether dynamics are local or global \citep{LocalDynamics}.
The dynamics that have appeared so far are from a global perspective,
and chaoticity represents the characteristic of the average of convergence or divergence around state transitions over a long period,
as seen in the definition of the Lyapunov exponent.
%The convergence or divergence varies depending on the state.
%It is often determined by the Lyapunov exponent,
%which indicates whether the neighborhoods around the observed trajectories converge or diverge on average over a long period.
Assuming that the system is dynamic with deterministic state transitions that can be described by non-linear differential equations.
The instantaneous convergence or divergence of a neighborhood of a given state
-- what is called `local dynamics'  -- is determined by the spectrum (or eigenvalues) of the Jacobian matrix
of the differential equation and varies with state transitions if the system is nonlinear. 
It is not a scalar nor a constant matrix, but each element of the matrix can be a function of the state.
Therefore, as shown in Fig.~\ref{fig:LocalDynamics}(A), the local dynamics in the state space can converge around one state
and diverge around another even in the same system.
Moreover, as shown in Fig.~\ref{fig:LocalDynamics}(B), even in the neighboring region of the same state,
the flow from a point displaced in one direction from the state can converge,
while that from a point displaced in another direction can diverge
if the Jacobian matrix has both positive and negative eigenvalues.

The author believes that humans are able to harmonize rationality and novelty
making full use of the degrees of freedom (DOFs) in the dynamics through learning,
which enables humans to sometimes step beyond common sense while thinking rationally.
However, the temperature parameter is often scheduled manually but is an unlearned scalar constant
and is not enough to change the randomness flexibly depending on the state or direction.
%Its change is sometimes scheduled manually but is not learned.

\newpage
\subsection{Introduction of Dynamic Reinforcement Learning (Dynamic RL)}
Comprehensively considering the above with the hypothesis in mind,
the author introduces a novel RL framework again
that significantly redefines the concepts of exploration and learning in an RNN, as follows.
\\
\begin{enumerate}
\item{An RL agent does not perform stochastic motion selection using external random noise.
The chaotic dynamics produced by the agent's RNN and the environment together embed exploration factors in its motion outputs.\label{Item:Chaos}}
\item{The dynamics are trained to enhance reproducibility when
the TD error is positive (i.e., the value is better than expected) and to enhance exploratory behavior when the TD error is negative
(i.e., the value is worse than expected) while maintaining chaotic dynamics. \label{Item:LearnDyn}}
\item{To achieve the objective (item \ref{Item:LearnDyn}), sensitivity \citep{Sensitivity} is used.
It is a local index that represents the degree of convergence or divergence of the input neighborhood
to the corresponding output neighborhood through the processing of each neuron.
\label{Item:UseSen}}
\end{enumerate}
The author's group already proposed the above hypothesis and item \ref{Item:Chaos} of the concept in \citep{IJCNN2015}.
%However, at that time, the recurrent connection weights could not be trained appropriately.
It was also demonstrated that chaotic dynamics operate as exploration without stochastic selection in actor-critic \citep{IJCNN2015},
reward-modulated Hebbian learning \citep{NN_Matsuki}, and TD3 \citep{TD3_Matsuki}
and that the ``edge of chaos'' was a favorable choice for network dynamics, leading to high learning performance in \citep{NN_Matsuki}.
% when weights were set
%so as that the system dynamics are around the edge of chaos \citep{NN_Matsuki}.%Furthermore, unlike the typical conventional learning,
%this learning does not aim to move the network output in a more appropriate direction at each moment,
%but rather aims at controlling the flow of the network state over time.
%The author will discuss it later\ref{Sec:Discussion}.

%However, considering their capabilities, the author believes it is likely that similar processes,
%if not all, are actually functioning in our brains.  (Transformer)

In Dynamic RL, an agent can explore using many DOFs because the motions with exploration are generated through its RNN
and can be learned.
The flexibility in exploration is expected to carry over directly into thinking through learning
while maintaining chaos in the global dynamics.
%Accordingly, unlike stochastic selection, learning local dynamics is expected to properly adjust divergence or convergence
%for each state and direction of deviation.
%Thus, that brings a huge degree of freedom in exploration, and turns it to be flexible.
% enables fine-tuned exploration
Furthermore, if the RNN architecture is hierarchical, learning is expected to control irregularity even in the abstract state space.
In this way, the author expects that a Dynamic RL agent, learning through high-DOF exploration,
will eventually be able to think about a variety of things, including abstract ones,
and sometimes generate completely new ideas on its own by flexibly utilizing both the irregular and the rational.

Currently, to improve learning efficiency or stability, many excellent RL methods,
such as PPO \citep{PPO}, SAC \citep{SAC}, A3C \citep{A3C}, TD3 \citep{TD3}, 
and experience replay with PER \citep{PER} or HER \citep{HER} have been used.
However, since Dynamic RL presents a newborn learning concept,
comparison targets are limited to simple conventional RL for a fair evaluation of basic performance.
%Notably, Transformer \citep{Transformer} has become dominant in AI,
%and is applied also in RL, such as Decision Transformer \citep{DecisionTransformer}.
%It is entirely distinct from conventional RL, and its function emergence is not yet well understood.
%%However, the emergence of functions inside has not been so clear yet.
%Given the remarkable performance of Transformer-based generative AI, a comparison or integration with it
%will likely become inevitable.
%However, since it differs significantly from Dynamics RL,
%the author prefers to set it aside for now to avoid confusion.

This paper first introduces Dynamic RL
and then examines whether this entirely new type of RL functions effectively as reinforcement learning
in a simple memory-required task and a dynamic pattern generation task, compared with conventional RL.

\section{Learning Method}
As a base architecture for RL, the actor-critic model is used,
in which the actor outputs do not represent probabilities for actions
but instead represent continuous motor commands.
Dynamic RL is applied solely to the actor, while the critic is trained by conventional RL using BPTT \citep{PDP}
although ideally, all learning should be dynamic.
For clarity, each actor and critic consists of a separate RNN with sensor signals as inputs.
%Q-learning is more widely used.
%However, it is the learning for discrete actions, and some more process is required
%before getting the final motor commands.
%From the view of building autonomous learning agents,
%there remains the problem how the process is acquired through RL.
%On the other hand, the outputs of the actor in actor-critic can be dealt with as continuous motion signals.

%Figure \ref{fig:neuron_forward} shows a general static-type neuron model with $m$ inputs.
In each dynamic neuron, its internal state $u$ at time $t$ is derived
as the first-order lag of the inner product of the connection weight vector ${\bf w}=(w_1, ... , w_m)^\mathrm{T}$
and input vector ${\bf x}_t=(x_{1t}, ... , x_{mt})^\mathrm{T}$ where $m$ is the number of inputs as
\begin{equation}
u_t = \left(1-\frac{\Delta t}{\tau}\right)u_{t-1}+\frac{\Delta t}{\tau}{\bf w}\cdot{\bf x}_t
\label{Eq:internal_state}
\end{equation}
where $\tau$ is a time constant and $\Delta t$ is the step width, which is 1.0 in this paper.
For static-type neurons, the internal state $u$ is just the inner product as
\begin{equation}
u_t = {\bf w}\cdot{\bf x}_t.
\label{Eq:internal_state_static}
\end{equation}
%by setting $\tau=\Delta t$.
The inputs ${\bf x}_t$ can be the external inputs or the pre-synaptic neuron outputs at time $t$,
%which may be outputs of neurons.
but for the feedback connections, where the inputs come from the same or an upper layer,
they are the outputs of the pre-synaptic neuron at time $t-1$. 
The output $o_t$ is derived from the internal state $u_t$ as
\begin{equation}
o_t = f(U_t)=f(u_t+\theta)
\label{Eq:output}
\end{equation}
where $U_t=u_t+\theta$, $\theta$ is the bias, and $f(\cdot)$ is an activation function,
which is a hyperbolic tangent in this paper.

Dynamic RL controls the dynamics of the system, including RNN, directly by adjusting the sensitivity \citep{Sensitivity} in each neuron.
Sensitivity is an index for each neuron that is the Euclidian norm of the output gradient
with respect to the input vector ${\bf x}$.
It is defined as
%how a neuron is sensitive to a small change in its inputs.
%It is defined as the Euclidean norm of the output gradient with respect to the input vector ${\bf x}$ as
\begin{equation}
s(U; {\bf w}) = \|\nabla_{\bf x} o\| = f'(U)\|{\bf w}\|.
\label{Eq:sensitivity}
\end{equation}
Here, $\| {\bf v} \| = \sqrt{\sum_i^mv_i^2}$ for a vector ${\bf v}=(v_1, ..., v_m)^\mathrm{T}$.
In the form of a vector elements, the sensitivity is represented as
\begin{equation}
s(U; {\bf w}) = \sqrt{\sum_i^m \left( \frac{\partial o}{\partial x_i} \right)^2} = f'(U)\sqrt{\sum_i^m w_i^2}\ .
\label{Eq:sensitivity_non_vector}
\end{equation}
Sensitivity refers to the maximum ratio of the absolute value of the output deviation $do$
to the magnitude of the infinitesimal variation $d{\bf x}$ in the input vector space.
It represents the degree of contraction or expansion from the neighborhood around the current inputs
to the corresponding neighborhood around the current output through the neuron's processing.
In the previous work \citep{Sensitivity}, it was defined only for static-type neurons
(Eq.~(\ref{Eq:internal_state_static})).
In this study, the same definition is also applied to dynamic neurons (Eq.~(\ref{Eq:internal_state})),
assuming that the infinitesimal variation $d{\bf x}$ of the input ${\bf x}$
changes slowly enough compared to the time constant $\tau$.
%it is assumed that the infinitesimal deviation $d{\bf x}$ of the input ${\bf x}$
%is a constant vector near the time $t$.
%By solving the linear asymptotic equation as in Eq.~(\ref{Eq:internal_state}),
%the deviation $du$ of the internal state can be represented as
%\begin{equation}
%du_t \approx \left\{1+\left(1-\alpha\right)+\left(1-\alpha\right)^2+...\right\}\alpha{\bf w}\cdot d{\bf x}_t 
%= \sum_{i=0}^\infty (1-\alpha)^i\alpha{\bf w}\cdot d{\bf x}_t = {\bf w}\cdot d{\bf x}_t
%\end{equation}
%where $0.0 < \alpha = \frac{\Delta t}{\tau} \leq 1.0$.
%Then the gradient of the internal state $u$ with respect to the input ${\bf x}$ becomes
%\begin{equation}
% \|\nabla_{{\bf x}_t} u_t\| = \|{\bf w}\|,
%\end{equation}
%and we can derive Eq.~(\ref{Eq:sensitivity}) as well also for the dynamic neurons.
%if the activation function $f$ is a monotonically increasing function.

In the previous research \citep{Sensitivity},  the author's group proposed sensitivity adjustment learning (SAL).
SAL was applied to ensure the sensitivity of each neuron in parallel with gradient-based supervised learning.
This approach is beneficial not only for maintaining sensitivity during forward computation in the neural network
but also for avoiding diverging or vanishing gradients during backward computation.
Because Dynamic RL incorporates SAL and sensitivity-controlled RL (SRL), which is an extension of SAL for RL,
SAL will be explained first.

In SAL, the moving average of sensitivity $\bar{s}$ is computed first as
\begin{equation}
 \bar{s}_t \leftarrow (1-\alpha) \bar{s}_{t-1}  + \alpha s_t
 \label{Eq:Ave_sen}
\end{equation}
where $\alpha$ is a small constant, and this computation is performed across episodes.
When the average sensitivity $\bar{s}$ is below a predetermined constant $s_{th}$,
the weights and bias in each neuron are updated locally to the gradient direction of the sensitivity as
\begin{align}
\Delta {\bf w}_t &= \eta_{SAL}\frac{\Delta t}{\tau} \nabla_{\bf w} s(U_t; {\bf w})
%                        = \eta_{SAL}\frac{\Delta t}{\tau} \nabla_{\bf w} \{f'(U_t)\|{\bf w}\|\}
                        = \eta_{SRL}\frac{\Delta t}{\tau} \left( f'(U_t)\frac{\bf w}{\| {\bf w} \|} + \| {\bf w} \| \nabla_{\bf w} f'(U_t) \right)
\label{Eq:SAL_ORG}\\
%\end{equation}
%and
%\begin{equation}
\Delta {\theta}_t &= \eta_{SAL} \frac{\Delta t}{\tau}\frac{\partial s(U_t; {\bf w})}{\partial \theta}
%                          = \eta_{SAL} \frac{\Delta t}{\tau}\frac{\partial \{f'(U_t)\|{\bf w}\|\}}{\partial \theta}
                          = \eta_{SRL}\frac{\Delta t}{\tau} \| {\bf w} \| \frac{\partial f'(U_t)}{\partial \theta}.
\label{Eq:SAL_Bias_ORG}
\end{align}
where $\eta_{SAL}$ is the learning rate for SAL.
$\Delta t / \tau$ is multiplied to adjust the update to the neuron's time scale.
By expanding the equation with the activation function being hyperbolic tangent,
\begin{align}
\Delta {\bf w}_t &= \eta_{SAL}\frac{\Delta t}{\tau} (1-o_t^2) \left( \frac{\bf w}{\| {\bf w}\|} - 2o_t\|{\bf w}\|{\bf x}_t \right)
\label{Eq:SAL} \\
%\end{equation}
%\begin{equation}
\Delta {\theta}_t &= -2 \eta_{SAL}\frac{\Delta t}{\tau} o_t(1-o_t^2) ||{\bf w}||
\label{Eq:SAL_Bias}
\end{align}
are derived.
%, where
%\begin{equation}
%\frac{do}{dU} = f'(U) = \frac{dtanh(U)}{dU} =\frac{1}{\cosh^2(U)} = 1- o^2.
%\end{equation}

\begin{figure}[t]
\centerline{\includegraphics[scale=0.35]{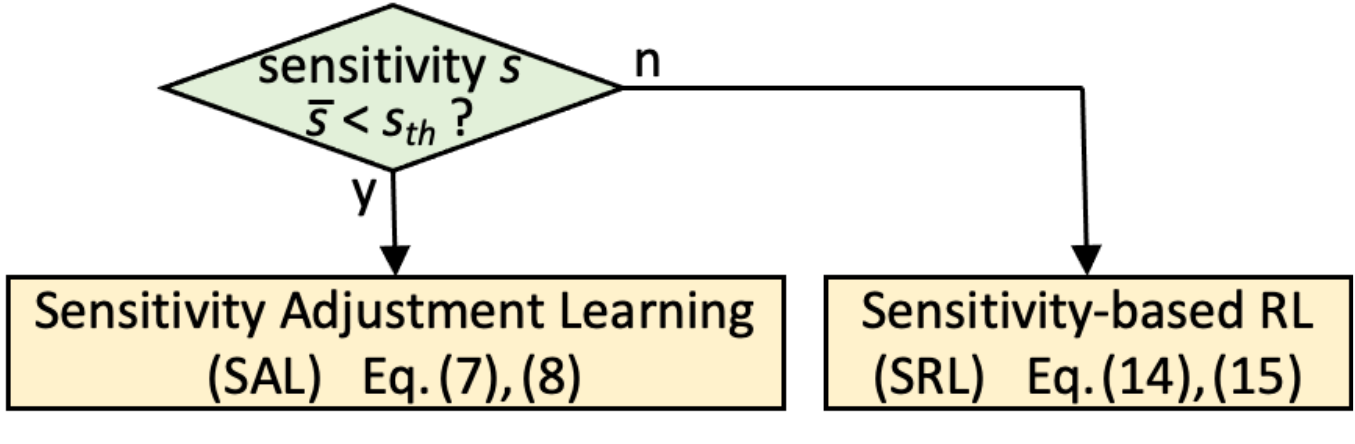}}
%\centerline{\includegraphics[scale=0.5, pagebox=cropbox, clip]{Task1.pdf}} 
\caption{Dynamic RL applies either SAL or SRL depending on the condition in each neuron.}
\label{fig:DynamicRL}
\end{figure}
\begin{figure}[ht]
\centerline{\includegraphics[scale=0.28]{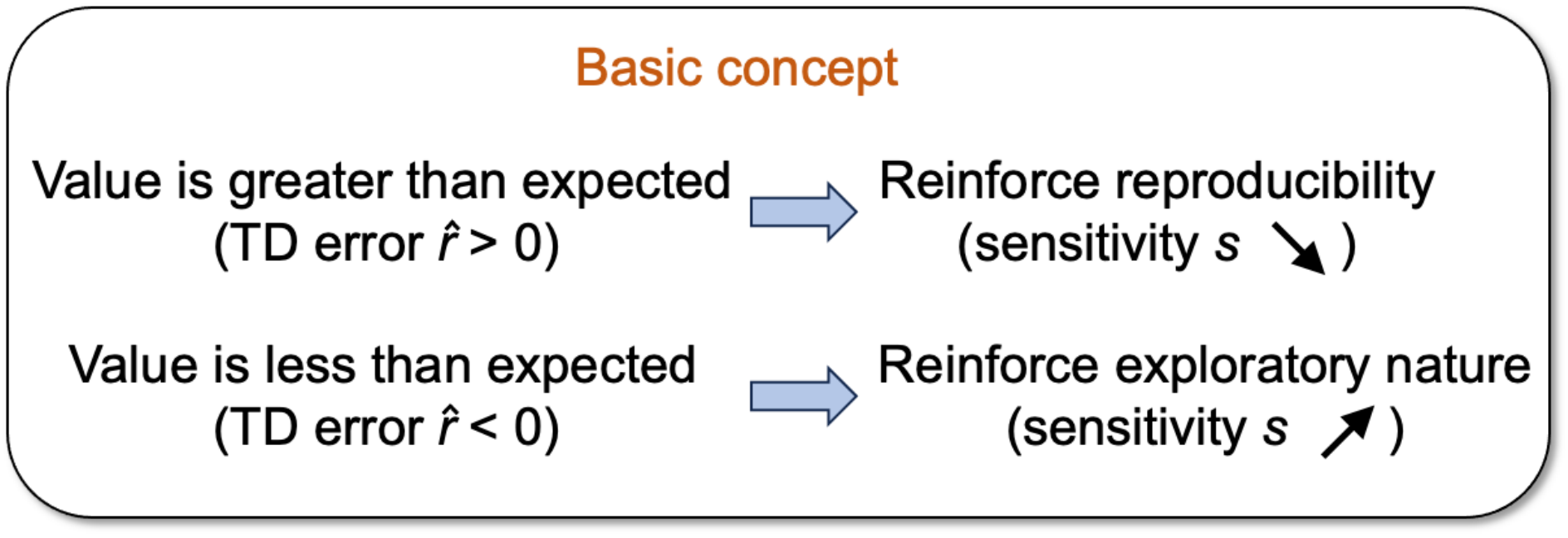}}
%\centerline{\includegraphics[scale=0.5, pagebox=cropbox, clip]{Task1.pdf}} 
\caption{Basic concept of Dynamic RL (or more specifically, SRL) proposed in this paper.}
\label{fig:BasicConcept}
\end{figure}

In Dynamic RL proposed here, as shown in Fig.\ref{fig:DynamicRL},
SAL is applied when the moving average of the sensitivity $\overline{s}$
is less than a constant $s_{th}$, otherwise sensitivity-controlled RL (SRL) is applied in each neuron.
%When not less
SAL always tries to increase the sensitivity in each neuron,
but whether SRL tries to increase or decrease the sensitivity depends on the temporal difference (TD) error ${\hat r}$ as
\begin{align}
\Delta {\bf w}_t &= -\eta_{SRL}\frac{\Delta t}{\tau} \hat{r}_t \nabla_{\bf w} s(U_t; {\bf w})\\
%\label{Eq:SRL} \\
%\end{equation}
%\begin{equation}
\Delta \theta_t &= -\eta_{SRL}\frac{\Delta t}{\tau} \hat{r}_t \frac{\partial s(U_t; {\bf w})}{\partial \theta}
%\label{Eq:SRL_Bias}
\end{align}
where $\eta_{SRL}$ is the learning rate for SRL.
TD error is computed as
\begin{equation}
\hat{r}_t = \gamma C_{t+1} + r_{t+1} - C_t = \gamma\left(C_{t+1}-\frac{C_t-r_{t+1}}{\gamma}\right)
\label{Eq:TDerr}
\end{equation}
where $\gamma\ (0.0<\gamma<1.0)$ is the discount factor, $C_t$ is the critic output (state value),
and $r_t$ is the reinforcement signal, which can be a reward or a penalty, at time $t$.
As the basic concept summarized in Fig.\ref{fig:BasicConcept},
when TD error is positive, in other words, the new critic (state value) $C_{t+1}$ is greater than the expected value
$\frac{C_t-r_{t+1}}{\gamma}$,
RL reduces the sensitivity to reinforce the reproducibility.
When it is negative, i.e., the new state value is less than expected,
RL makes the sensitivity greater to reinforce the exploratory nature.
This is expected to control the local convergence or divergence, depending on how good or bad the state is.

Upon expansion, we obtain,
\begin{align}
\Delta {\bf w}_t &= -\eta_{SRL}\frac{\Delta t}{\tau} \hat{r}_t \left( f'(U_t)\frac{\bf w}{\| {\bf w} \|} + \| {\bf w} \| \nabla_{\bf w} f'(U_t) \right)\\
\Delta \theta_t &= -\eta_{SRL}\frac{\Delta t}{\tau} \hat{r}_t \| {\bf w} \| \frac{\partial f'(U_t)}{\partial \theta}.
\end{align}
By further expanding as the activation function $f(\cdot)$ being $\tanh$,
\begin{align}
\Delta {\bf w}_t &= - \eta_{SRL}\frac{\Delta t}{\tau} \hat{r}_t (1-o_t^2) \left( \frac{\bf w}{\| {\bf w}\|} - 2o_t\|{\bf w}\|{\bf x}_t \right)
\label{Eq:SRL}\\
\Delta \theta_t &= 2\eta_{SRL}\frac{\Delta t}{\tau} \hat{r}_t o_t (1-o_t^2) \|{\bf w}\|.
\label{Eq:SRL_Bias}
\end{align}
%The equation is rewritten in the element form as
%\begin{equation}
%\Delta w_i = \eta_{SAL} \frac{(1-o^2) \left\{ w_i -2ox_i \sum_k w_k^2 \right\}}{\sqrt{\sum_k w_k^2}}.
%\end{equation}
%The author calls the first term $-\eta \hat{r} (1-o^2){\bf w}/\left|{\bf w}\right|$ the linear term,
%which is originated from $|{\bf w}|$ in Eq.~(\ref{Eq:sensitivity}).
%The second term $2 \eta \hat{r} (1-o^2) o|{\bf w}|{\bf x}$ is called non-linear term,
%which is originated from $f'(x)$ in Eq.~(\ref{Eq:sensitivity}).
%Different from the case of weight, bias $\theta$ cannot increase the sensitivity directly, but
%can increase it indirectly by updating the bias so that the value $U$ becomes closer to 0.0.
Notably, this computation can be done locally in each neuron except for receiving the TD errors.
Furthermore, since the dynamics are generated not only by the loops inside the RNN
but also influenced by the loops that are formed with the outside world,
this learning can be applied to all the neurons, including those outside the loop in the RNN, including the output neurons.

In the following simulations, the proposed RL is compared to the conventional RL using BPTT.
%Then, the conventional RL used here is explained next.
Now many techniques have been proposed to improve the performance, but for a pure comparison of base methods,
simple learning using gradient-based BPTT is employed.
In Dynamic RL, the motor command vector ${\bf M}_t$
is a function ${\bf M}(\cdot)$ of the actor output vector ${\bf A}_t$ as ${\bf M}_t = {\bf M}({\bf A}_t)$,
%is identical to the actor output vector ${\bf A}_t$,
but in the conventional RL, since a random noise vector ${\bf \epsilon}_t$ is added to the actor output vector
as explorations, the actual motor command vector ${\bf M}_t$ is expressed as
\begin{equation}
%{\bf M}_t = {\bf A}_t + {\bf \epsilon}_t
{\bf M}_t = {\bf M}({\bf A}_t + {\bf \epsilon}_t)
\end{equation}
For conventional RL, training signals for the actor network are derived as
\begin{equation}
{\bf A}_{train,t} = {\bf A}_t + \hat{r}_t {\bf \epsilon}_t .
\label{Eq:ConvRL}
\end{equation}
Then, the actor network is trained based on the BPTT method by these training signals.
In this paper, it learned 10 or 20 steps backward in time, depending on the task.
While, in the Dynamic RL, since no calculation going back through time is necessary,
the computational cost is considerably smaller than in the case of conventional RL.

\begin{figure}[t]
\centerline{\includegraphics[scale=0.31]{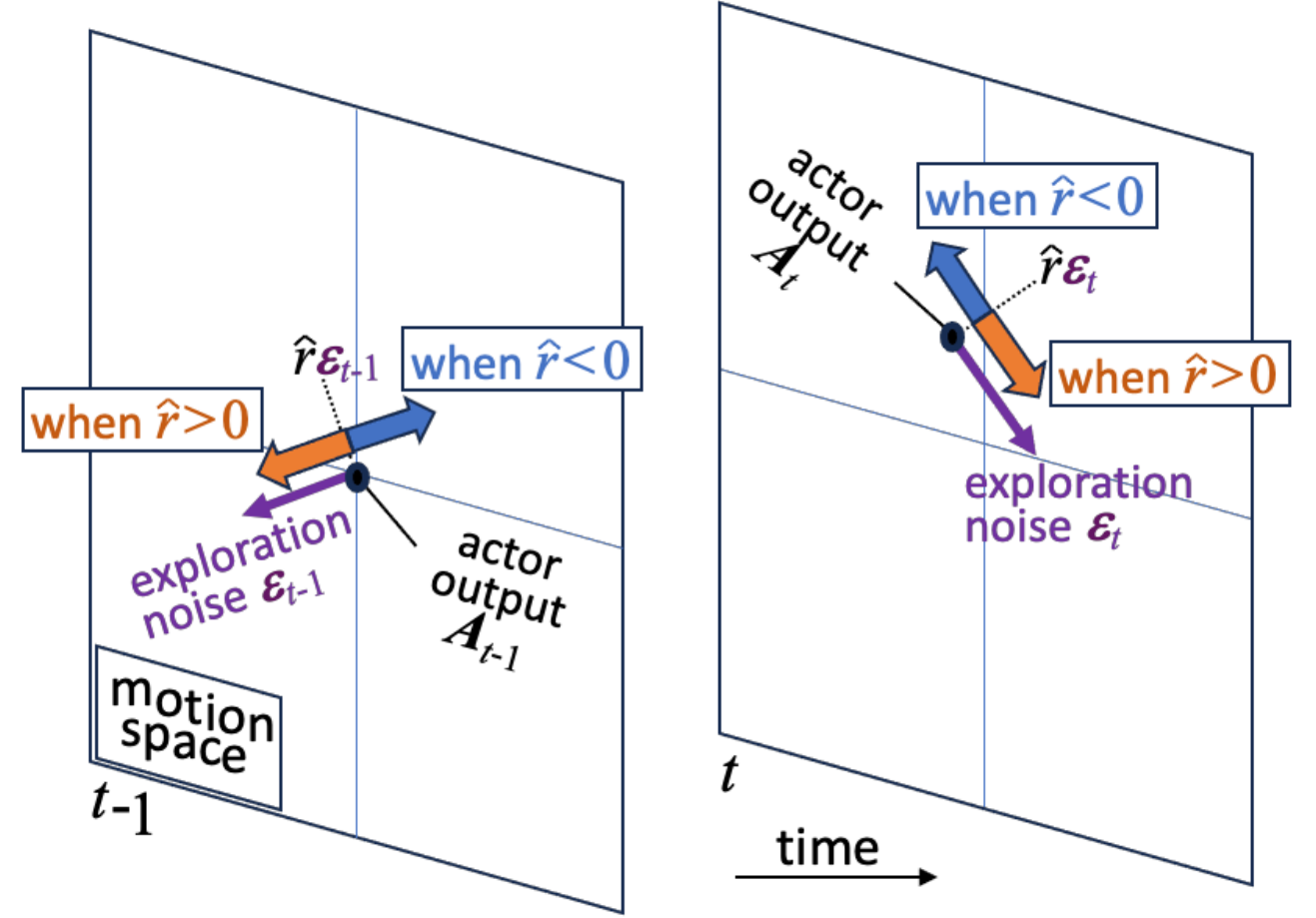}}
%\centerline{\includegraphics[scale=0.5, pagebox=cropbox, clip]{Task1.pdf}} 
\caption{A conceptual diagram of conventional RL.
RL aims to control the actor output vector based on the TD error.
It does not utilize information about time changes in the RNN's state and is closed only at each step.}
\label{fig:ConvRL}
\vspace{5mm}
\centerline{\includegraphics[scale=0.31]{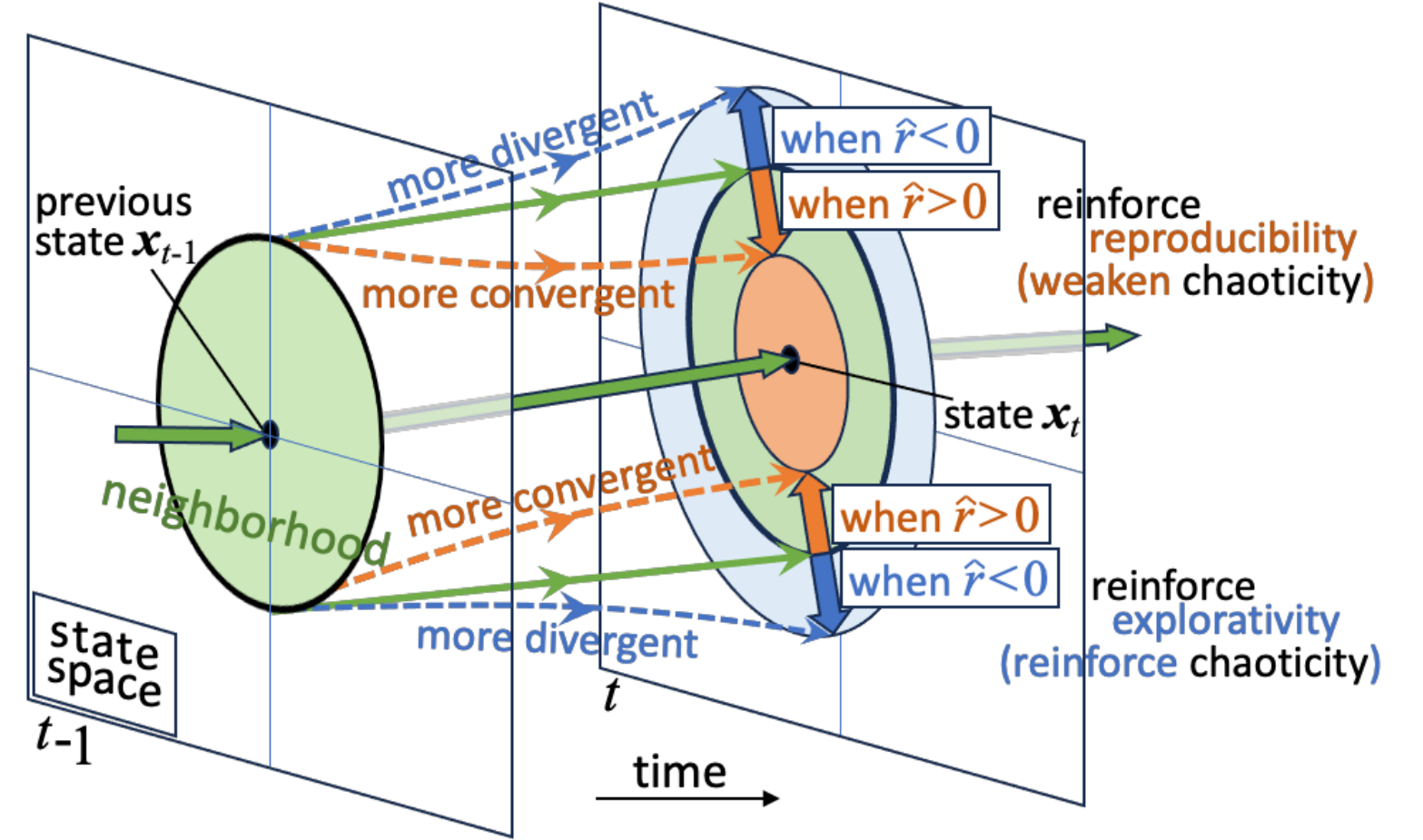}}
%\centerline{\includegraphics[scale=0.5, pagebox=cropbox, clip]{Task1.pdf}} 
\caption{A conceptual diagram of Dynamic Reinforcement Learning (RL).
RL aims to control the convergence or divergence of the flow around state transitions according to the TD error
by controlling the sensitivity in each neuron.}
\label{fig:DYN_RL}
\end{figure}
%As described in the Introduction, 
Dynamic RL has a significant difference in the way of learning
from conventional RL.
For better understanding, the author attempts to illustrate their differences with diagrams at the expense of accuracy.
In the conventional RL, external noise ${\bf \epsilon}$  is added to the actor output vector ${\bf A}$.
As shown in Fig.~\ref{fig:ConvRL}, according to Eq.~(\ref{Eq:ConvRL}),
if the value function is better than expected, i.e., if the TD error $\hat{r}$ is positive,
the network is trained to move the output vector ${\bf A}$ to the direction of the noise ${\bf \epsilon}$.
By contrast, if the TD error $\hat{r}$ is negative, the network is trained to move the output vector ${\bf A}$ to the opposite direction.
This RL does not use the temporal change in the outputs or network states; rather, it considers only the outputs at that moment in time.
All the weights and biases are updated to move the output vector with the help of the gradient method
using error backpropagation even through time.

On the other hand, Dynamic RL does not aim to move the state or output directly,
but as shown in Fig.~\ref{fig:DYN_RL}, it aims to control the convergence or divergence of the neighborhood
around the state transition by changing each neuron's sensitivity depending on the TD error.
The concept of controlling dynamics can also be applied to other types of learning, such as supervised learning.
The author refers to it as Dynamic Learning from a broader perspective and will discuss it in the subsection
\ref{subsec:Future}.
%Therefore, the learning in the neurons that are not included in any loop in the RNN
%also influences the dynamics.

%In this paper, to improve the performance further,
%another gradient-based learning is applied to the output neurons referring to \citep{Hoerzer,Matsuki}.
%Here, the deviation from the moving average is computed.
%\begin{equation}
%\tilde{o}_t = {o}_t - \bar{o}_t
%\end{equation}
%where $\bar{o}_t = 0.8 \bar{o}_{t-1} + 0.2 {o}_t$, and the weight vector is updated
%by the product of it and TD error as
%\begin{equation}
%\Delta {\bf w}_t = \eta_{grad} \hat{r}_t \tilde{o}_t {\bf x}_t.
%\label{Eq:GradL}
%\end{equation}
%The biases are updated as
%\begin{equation}
%\Delta {\bf \theta}_t = \eta_{grad} \hat{r}_t \tilde{o}_t.
%\label{Eq:GradL_Bias}
%\end{equation}
This concept should also be introduced to the critic network,
but here, conventional learning is used for the critic, regardless of how the actor network is trained.
The training signal is derived as
\begin{equation}
C_{train,t} = \gamma C_{t+1} + r_{t+1} = C_t + \hat{r}_t,
\label{Eq:C_train}
\end{equation}
and the critic network is always trained with BPTT using this training signal.

In Dynamic RL, the network outputs were often saturated (close to $1.0$ or $-1.0$ in hyperbolic tangent),
and it is difficult to perform fine and smooth control.
To avoid saturation, the regularization was applied only to the output layer's connection weights in the actor network as
\begin{equation}
  \Delta {\bf W} = -\eta_{reg} {\bf W}.
  \label{Eq:Regularize}
\end{equation} 
This learning was applied in both Dynamic and conventional RL cases.
% for fair comparison.
 
Furthermore, one more technique used in this paper is ``critic raising''.
When an agent cannot reach its goal for a long time,
since the critic output becomes small, the gradient of the critic also becomes small.
Therefore, referring to the ``optimistic initial value'' \citep{Sutton1998},
when the moving average $\bar{C}$ of the critic output $C$ is less than a constant $C_{th}$,
the bias of the output layer in the critic network is increased to raise the critic value as
\begin{equation}
  \Delta \theta_t = \eta_{raise} (C_{th}-\bar{C}_t)
  \label{Eq:Raise_Critic}
\end{equation} 
where $\bar{C}$ is the moving average of the critic output $C$ as 
\begin{equation}
 \bar{C}_t \leftarrow (1-\beta) \bar{C}_{t-1}  + \beta C_t
 \label{Eq:Ave_C}
\end{equation}
where $\beta$ is a small constant, and this computation is performed across episodes.
%, but except for the preparation steps,
%in which only the RNN was computed without actually moving for preparation.
%This was also applied in both Dynamic and conventional RL cases.

\section{Simulations}
To examine the learning ability of Dynamic RL, the author applied it to two dynamic tasks:
a sequential navigation task, which served as a memory-required task, and a slider-crank control task,
which served as a dynamic pattern generation task.
%In both tasks, the results of the conventional RL in which BPTT was used is shown for reference.
%but because the proposed method has a large space to be improved,
%strict comparison was not done,

The network architecture used is shown in the left figure in Fig.~\ref{fig:Task_SW},
regardless of the learning method or task, although the parameters used are different.
The architecture consists of an actor network and a critic network;
as mentioned, Dynamic RL was applied only to the actor network here.
The critic network was trained in the same manner as conventional RL.
The critic network was an Elman-type RNN with a single hidden layer comprising 50 neurons, having feedback connections to itself.
It was trained to minimize the square of the TD-error using BPTT
with the training signal as in Eq.~(\ref{Eq:C_train}).

The actor network used a multiple timescale RNN (MTRNN) \citep{MTRNN}
to improve performance and verify whether the proposed RL would function
on a more complex architecture than simple Elman-type RNNs.
The MTRNN comprised two hidden layers with different time constants and one output layer.
The two hidden layers were interconnected, and one of them, called upper hidden layer (100 neurons),
did not have any connection to the input or output layer; however, it had self-feedback connections.
Although the other hidden layer, called lower hidden layer (200 neurons), had connections with the input and output layers,
it did not have self-feedback connections.
Neurons in the upper hidden layer in the actor network were dynamic,
while the rest, including all neurons in the critic network, were static (time constant $\tau=1$).
All connected layers are fully connected in both networks.

%The number of neurons and the time constant are shown in the Table \ref{Table:Num_Neuron}
%\begin{table}[ht]
%\label{Table:Num_Neuron}
%\begin{center}
%\caption{Neral network architecture}
%    \vspace{2mm}\small
%  \begin{tabular}{c|c|c}
%   \toprule
%     \multicolumn{3}{c}{number of neurons}\\
%    \midrule
%     critic\_net & & 50\\
%    \multirow{2}{*}{actor\_net} & upper layer & 100\\
%    & lower layer & 200\\
%   \bottomrule
%   \toprule
%     \multicolumn{3}{c}{neuron model $\tau$}\\
%    \midrule
%     actor\_net & upper layer & dynamic ($\tau=5.0$)\\
%    \multicolumn{2}{c}{others} & static\\
%   \bottomrule
%   \end{tabular}
%\end{center}
%\end{table}

In the proposed Dynamic RL, either SRL or SAL
% (Eqs.(\ref{Eq:SRL},\ref{Eq:SRL_Bias}))  (Eqs.(\ref{Eq:SAL},\ref{Eq:SAL_Bias}))
was applied to each hidden neuron depending on its average sensitivity $\bar{s}$ as shown in Fig.~\ref{fig:DynamicRL};
meanwhile, SAL was not applied to each output neuron to avoid the risk that SAL would cause the output to deviate from the ideal value.
To see the difference from the conventional RL and also how SRL and SAL worked respectively,
six learning conditions varying the method of applying learning were compared in the following
as in Table \ref{Tab:CompSet}.
\begin{table}[t]
  \label{Table:LearningCases}
  \begin{center}
    \vspace{0.5cm}
    \caption{Six learning conditions varying the application methods of learning for the actor network compared in the following simulations.
    (DYN\_RL: Dynamic RL, Conv\_RL: conventional RL, Reg: regularization (Eq.~(\ref{Eq:Regularize})),
     SRL \& SAL: SRL (Eqs.~(\ref{Eq:SRL})(\ref{Eq:SRL_Bias})) and SAL (Eqs.~(\ref{Eq:SAL})(\ref{Eq:SAL_Bias})), 
     but actually one of them is applied depending on the condition as in 
     Fig.~\ref{fig:DynamicRL} at each step in each neuron, BPTT: Backpropagation Through Time, 
     BP: Backpropagation not going back in time.
   In convenience, the notation `BP' is used also for the output layer
   where no propagation of error signal is necessary.
   The training signals for `BPTT' or `BP' are as in Eq.~(\ref{Eq:ConvRL}).)}
    \label{Tab:CompSet}
    \vspace{5mm}
    \small
      \begin{tabular}{l|ccc}
        \toprule
        \multicolumn{1}{c|}{\multirow{2}{*}{Name}}
        & \multicolumn{3}{c}{combination(in: input, hid: hidden, out: output)}\\ \cline{2-4}
             & hid $\to$ out & hid $\to$ hid & in $\to$ hid\\
         \midrule
         (A) \ \ DYN\_RL (proposed) & SRL \& Reg & SRL \& SAL & SRL \& SAL\\
         (A-1) DYN\_RL& \multirow{2}{*}{SRL \& Reg} & \multirow{2}{*}{SAL} & \multirow{2}{*}{SRL \& SAL}\\
          \ \ \ \ \ \ (No SRL for hid$\rightarrow$hid)\\
         (A-2) DYN\_RL (No SAL) & SRL \& Reg & SRL & SRL\\
         \midrule
         (B) \ \ Conv\_RL (BPTT) & BP \& Reg & BPTT & BPTT\\
         (B-1) Conv\_RL (BPTT+ SAL) & BP \& Reg & BPTT \& SAL & BPTT \& SAL\\
         (B-2) Conv\_RL (BP+SAL) & BP \& Reg & BP \& SAL & BP \& SAL\\ 
    \bottomrule
    \end{tabular}
    \label{Table:CompSet}
  \end{center}
\end{table}

%\begin{table}[ht]
%  \begin{center}
%    \caption{Learning methods for the actor network compared in the following simulations.\\
%                   (``hid'': applying to all the hidden layers, ``out'': applying to the output layer)}
%    \vspace{1mm}\small
%      \begin{tabular}{l|cccc} \toprule
%        \multicolumn{1}{c}{Name}\vline
%             & \hspace{-0.2cm}\begin{tabular}{c}SRL\\Eq.(\ref{Eq:SRL})\end{tabular} \hspace{-0.2cm}
%             & \hspace{-0.2cm}\begin{tabular}{c}SAL\\Eq.(\ref{Eq:SAL})\end{tabular}\hspace{-0.2cm}
%             & \hspace{-0.4cm}\begin{tabular}{c}Grad\\Eq.(\ref{Eq:GradL},\ref{Eq:GradL_Bias})\end{tabular} 
%                 \hspace{-0.4cm}
%             & \hspace{-0.5cm}\begin{tabular}{c}BPTT or BP\\Eq.(\ref{Eq:ConvRL})\end{tabular}\hspace{-0.5cm}\\ 
%         \midrule
%         (A) DYN\_RL (proposed) & hid, out & hid & out & - \\
%         (A-1) DYN\_RL (SRL: out only)& out & hid & out & -\\
%         (A-2) DYN\_RL (No SAL) & hid, out & - & out & -\\
%         (A-3) DYN\_RL (No Grad) & hid, out & hid & - & - \\
%         \midrule
%         (B) Conv\_RL & - & - & - & \hspace{-0.2cm}hid, out\hspace{-0.2cm}\\
%         (B-1) Conv\_RL + SAL & - & hid & - & \hspace{-0.2cm}hid, out\hspace{-0.2cm}\\
%         (B-2) Conv\_RL (No BPTT)+SAL & - & hid & - & \hspace{-0.2cm}hid, out\hspace{-0.2cm}\\ 
%    \bottomrule
%    \end{tabular}
%    \label{Table:CompSet}
%  \end{center}
%\end{table}

\subsection{Sequential navigation (memory-required) task}
\begin{figure*}[th]
\centerline{\includegraphics[scale=0.27]{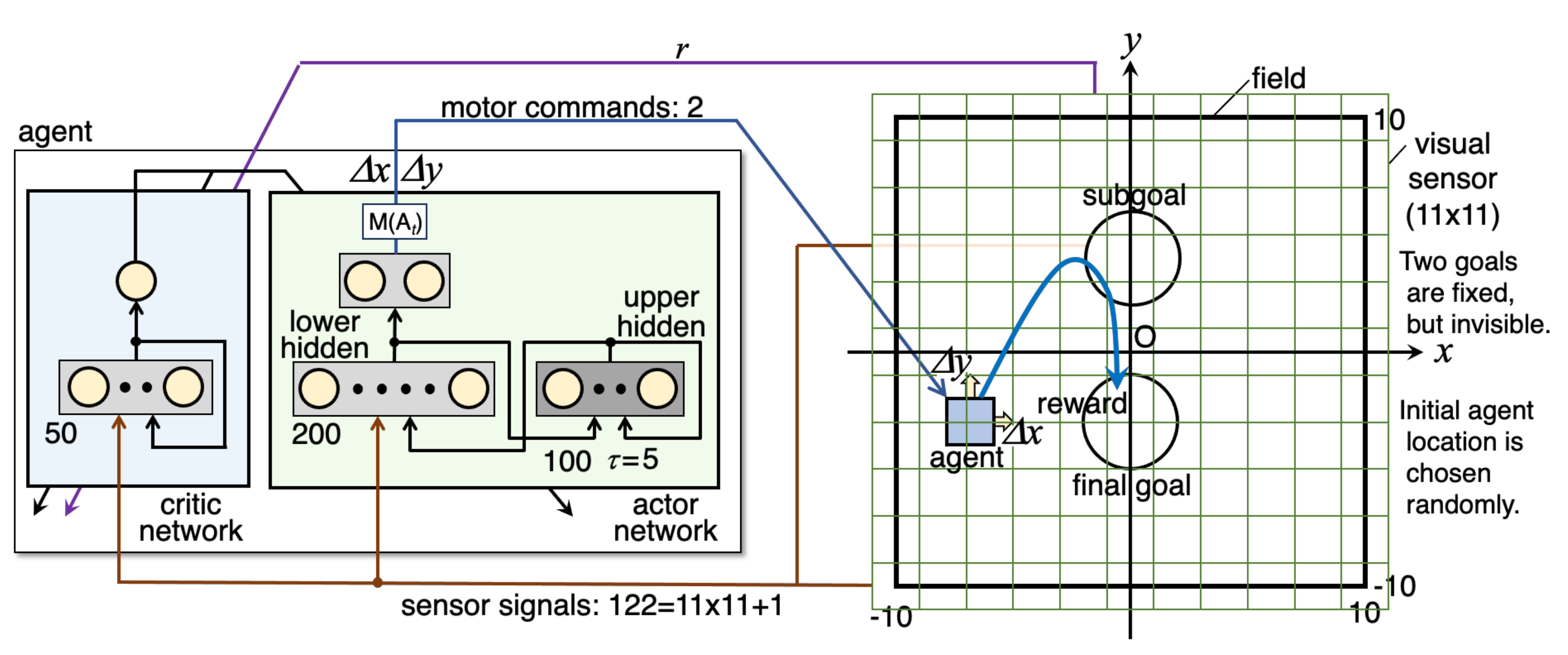}}
\caption{Sequential navigation task and the RNN used in the simulation.}
\label{fig:Task_SW}
\end{figure*}

In this simulation, as shown in the right figure of Fig.~\ref{fig:Task_SW},
there is a $20 \times 20$ size walled field centered at the origin.
An agent starts each episode at a randomly chosen location
weighted on the peripheral area of the field to accelerate learning (\ref{App:Task}).
% In order to promote the learning in the peripheral area, once per every 4 episodes,
% the location is limited to a point on the four sides of the $18\times 18$ square
% whose center is on the origin.
The agent received a reward of $r=0.8$ ($r$: reinforcement signal)
when the center of the agent reached the final goal at $(0.0, -3.0)$ after reaching the subgoal at $(0.0, 4.0)$.
At every step when the agent center is on the final goal before reaching the subgoal,
it receives a small penalty of $r=-0.03$.
The shape of each goal (final goal or subgoal) is a circle with a radius of 2.0, 
and when the center of the agent comes within the circle, the agent is considered to have reached the goal.
There is a visual sensor with $11 \times 11=121$ visual cells, and only the agent appears on it.
Each sensor cell covers a $1 \times 1$ square, which is identical to the agent's size.
%the size of each is $1\times1$.
%The sensor is fixed with its center at the origin, and the cells are arranged in a square without overlap.
%The sensor captures only the agent that can be seen as a $1\times1$ square,
%and each cell outputs the area occupied by the agent in the range from 0 to 1.
The agent cannot know the location of each goal from the sensor signals.
The agent also receives one more sensor signal.
Only when the agent's center is within the subgoal, it is 2.0; otherwise, it is 0.0.
Therefore, the agent must memorize whether it has already reached the subgoal or not in the RNN
and switch its behavior according to the memory.

The agent receives 122 sensor signals in total and inputs them to both the critic and actor networks.
The actor network has two outputs, which are used as the agent's movement vector after normalizing
to make the movable area circular, as described in \ref{App:Task} at each step.
In each episode, the agent resets all the internal states of the two RNNs to 0.0 at first,
and for preparation, it computes the RNNs for the first five steps without moving or counting steps.
%after being multiplied by 1.25 and clipped between the values of -1.0 and 1.0.
%Furthermore, the actual one-step movement of the agent is normalized while maintaining its direction
%so that its range of movement is confined to a circle with a radius of 1.0.
%For example, if the actor outputs are (0.0, 0.9), the agent will move by (0.0, 1.0),
%and if they are (0.3, 0.4), it will move by $(0.3, 0.4)$.
When the agent reaches a wall around the field, it stops there, but no penalty is given.
The episode terminates either when the agent receives a reward or after 200 steps have passed.
At each step, the agent updates all the weights and biases in the critic network
using the training signal as Eq.~(\ref{Eq:C_train}),
and those in the actor network according to Eqs.~(\ref{Eq:SRL})(\ref{Eq:SRL_Bias})(\ref{Eq:SAL})(\ref{Eq:SAL_Bias})(\ref{Eq:Regularize}).
When it received a reward, the critic at the next step $P_{t+1}$ was set to 0.0.
The main used parameters are shown in Table \ref{Table:Parameters1}.
In Dynamic RL, the spectral radius of the initial self-feedback weight matrix of the upper hidden layer in the actor network
was as large as 3.0 to ensure that the system, including actor network, generates chaotic dynamics,
and the agent can explore the field.
It was 1.3 in the conventional RL from the viewpoint of appropriate error propagation.
% BPTT or BP was used in the actor network in the case of conventional RL, and was always used in the critic network.
%Detailed parameters are shown in \ref{App:Table}.
\begin{table}[t]
  \begin{center}
    \caption{Parameters used in the sequential navigation (memory-required) task. (MA: Moving Average)}
    \vspace{2mm}\small
      \begin{tabular}{c|c} \toprule
        \begin{tabular}{c}Number of Neurons\\ (input-hidden[lower-upper]-output)\end{tabular}
         &  \begin{tabular}{c}122-50-1 (critic)\\
        122-[200-100]-2 (actor) \end{tabular} \\ \midrule
               \begin{tabular}{c}Spectral Radius of Initial \\ Self-feedback Connection Weight Matrix\end{tabular}
        & 
        \begin{tabular}{c}1.3 (critic, actor(Conv\_RL))\\
        3.0 (actor(DYN\_RL)) \end{tabular} \\ \midrule
        Neuron Model	& 
        \begin{tabular}{c}
            Dynamic with $\tau=5\ \ \ \ \ \ \ \ \ \ \ \ \ \ $\\(upper hidden layer in actor net.)\\
            Static (others)
        \end{tabular}
        \\ \hline
	Target Sensitivity for SAL  $s_{th}$ in Fig.~\ref{fig:DynamicRL}& 1.3\\ \midrule
	Discount Factor $\gamma$ in Eqs.~(\ref{Eq:TDerr}),(\ref{Eq:C_train})& 0.98\\ \midrule
	Rate of Regulation $\eta_{reg}$ in Eq.~(\ref{Eq:Regularize}) & $1e^{-6}$\\ \midrule
	Raise Critic Criterion $C_{th}$ in Eq.~(\ref{Eq:Raise_Critic})  & 0.1\\ \midrule
	Rate of Raising Critic $\eta_{raise}$ in Eq.~(\ref{Eq:Raise_Critic}) & 0.0005\\ \midrule
%        learning rate $\eta_{SAL}$ in Eq.~(\ref{Eq:Delta_w}) & 0.00002\\ \hline
%	decay rate $\beta$ in Eq.~(\ref{Eq:decay}) & $ 0.99 $\\ \hline
	Const. $\alpha$ for MA of Sensitivities in Eq.~(\ref{Eq:Ave_sen}) &  0.001\\ \midrule
	Const. $\beta$ for MA of Critic in Eq.~(\ref{Eq:Ave_C}) &  0.0001\\ \midrule
	Truncated Number of Steps in BPTT & 20\\ \midrule
	Initial Weights and Biases, Learning Rate & Refer to \ref{App:Table}\\ \bottomrule
      \end{tabular}
    \label{Table:Parameters1}
  \end{center}
\end{table}

In one simulation run, the agent learned for 20,000 episodes first and then for another 20,000 episodes
after swapping the roles of the actor's two outputs to observe its adaptability to environmental changes.
For each learning condition, a total of 40 simulation runs with different random sequences were performed.
The sequences were used to determine the initial connection weights, biases, and also the initial agent location at each episode.
%When the same random sequence was used, the initial agent location in the same episode was the same
%regardless of the learning method.
If the reduction in the average number of steps to the final goal was too small despite some progress in learning,
the simulation run was treated as a ``failure'' and terminated midway through the learning process (refer to \ref{App:Task} for details).
After each of the 20,000 and 40,000 episodes, a test was performed.
The agent started from 46 initial locations, arranged in a grid pattern with a 3.0 unit interval,
ranging from $-9.0$ to $9.0$ on the field, excluding the two goal areas, and did not learn. 
The random noise for exploration was not added to the actor outputs even in the case of conventional RL.
If the average number of steps in either test exceeded 20, the simulation run was deemed ``overrun.''
A simulation run was considered a ``success'' when it was neither a ``failure'' nor ``overrun.''

\begin{figure}[t]
\centerline{\includegraphics[scale=0.7]{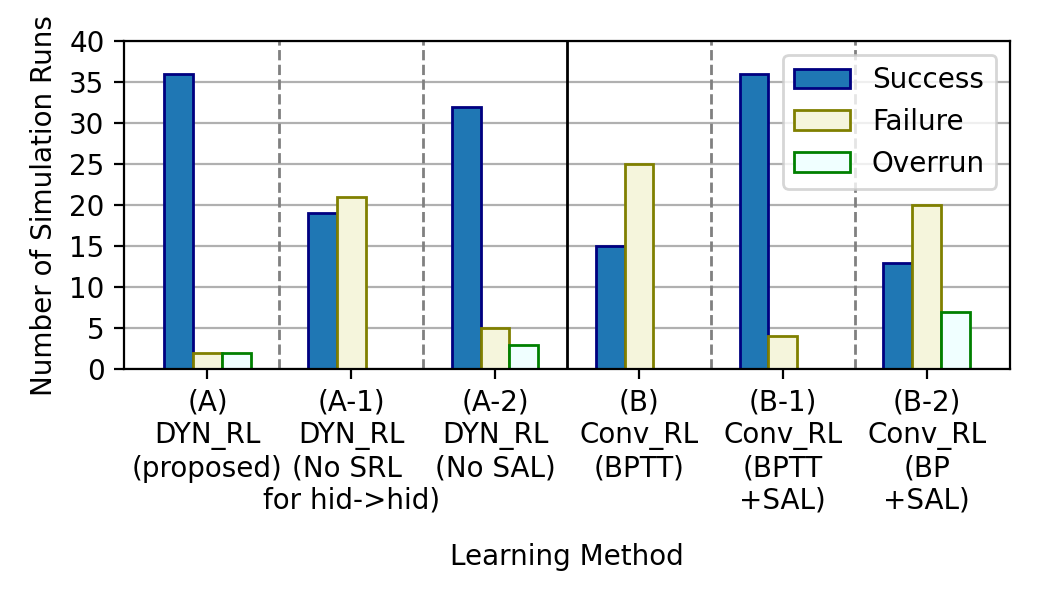}}
\caption{Comparison of success rate among six learning conditions as in Table~\ref{Table:CompSet}
in the sequential navigation task.}
\label{fig:CompPerf1_SW}
\end{figure}
First, Fig.~\ref{fig:CompPerf1_SW} compares the success rates from 40 simulation runs
across the six learning conditions listed in Table \ref{Table:CompSet}.
The (A) Dynamic RL proposed in this paper achieved 36 successes, 2 failures, and 2 overruns out of the 40 runs.
In the two failure runs, learning failed before the environmental change, and no failure occurred afterward.
In the two overruns, the average number of steps during learning was consistently below 20.0 just before the test;
however, due to some instability in Dynamic RL, it occasionally exceeded 20 steps in the test.

The success rate for (B) conventional RL using BPTT is only about 40\%, which is lower than for (A) Dynamic RL.
However, in the case of (B-1), where SAL was applied to the hidden neurons in parallel with conventional RL,
the results improved to 36-4-0, making them comparable to those of case (A) Dynamic RL.
Therefore, the case (B-1) will serve as the main comparison target for Dynamic RL hereafter.
When the error was not propagated to the past in the case of (B-2),  
the performance was worse than in the case of (B-1).

Meanwhile, comparing (A), (A-1), and (A-2) suggests that both SRL and SAL are necessary
for good performance in Dynamic RL.
Different from the case of the chaos-based RL method from the previous work \citep{IJCNN2015},
applying SRL to the weights between hidden neurons clearly improved the results.
%It can be said that Dynamic RL proposed here is effective also to the learning of weights among hidden neurons.
%However, though BPTT was applied, it is decreased to around 50\% if learning was not done for the past as (g).
%In (A) Dynamic RL, in which SAL was applied to the hidden layers by default,
%the agent sometimes took more than 20 steps on average in the tests,
%and the success rate was slightly lower than (B-1) conventional RL.
%but when SAL was not applied, the learning was almost unsuccessful as (A-2).
%The first thing that can be said from the results indicates at first that, 
%SAL, which can ensure sensitivity in each neuron, plays a significant role for the successful learning.
%Then it can be seen that SRL works for the learning of memory-required task.
%When SRL was applied only for the output layer,
%the success ratio was not so small, but as mentioned.

\begin{figure}[ht]
\centerline{\includegraphics[scale=0.6]{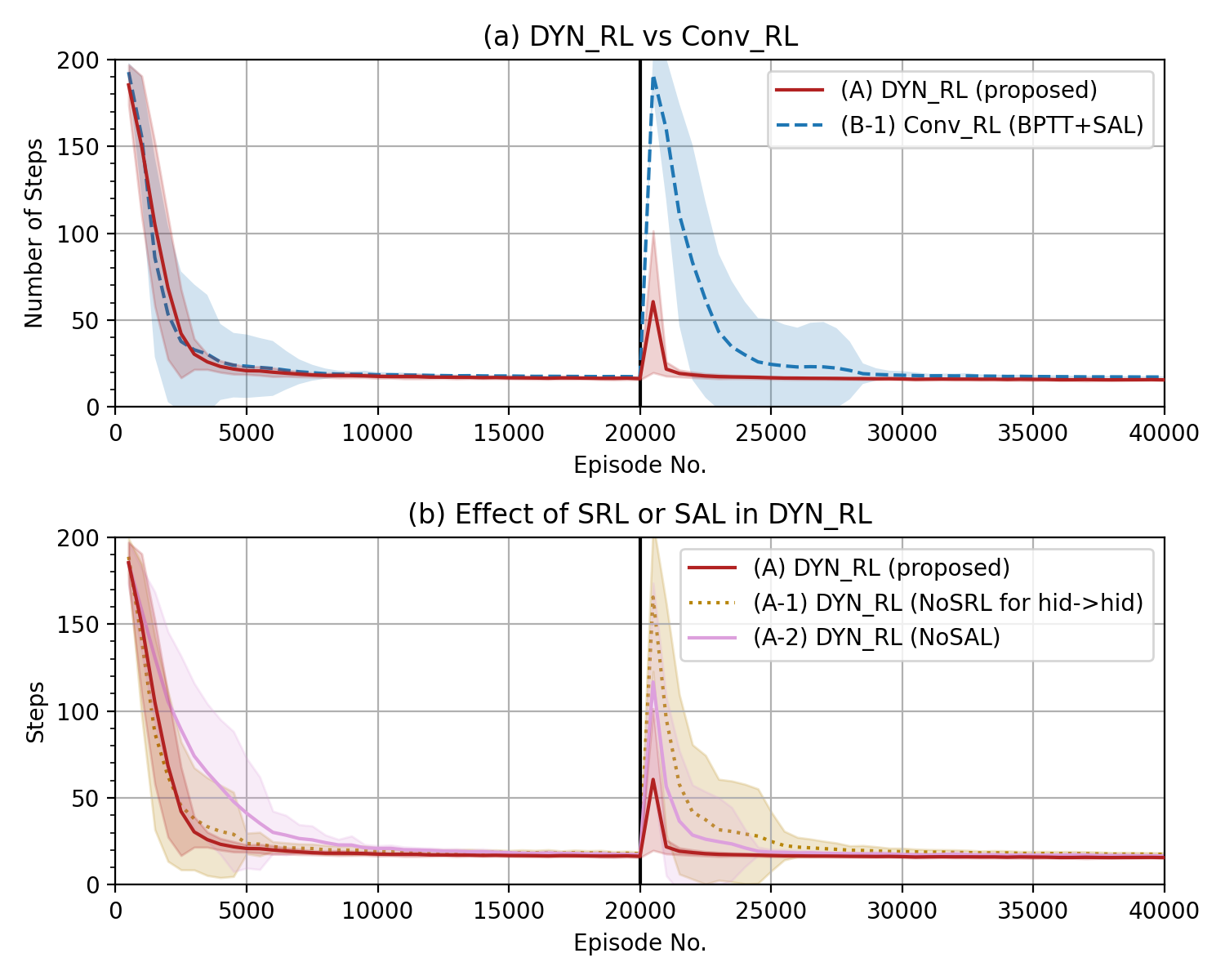}}
\caption{Learning curve for the sequential navigation task. The vertical axis indicates the average number of steps
necessary to reach the final goal successfully over 500 episodes. The plots and shaded areas indicate the mean and standard deviation
over successful simulation runs.
(a) shows the comparison result between Dynamic RL and conventional RL.
(b) shows the comparison result among three cases to assess the effect of SRL or SAL.}
\label{fig:LearningCurve_SW}
\end{figure}
Figure \ref{fig:LearningCurve_SW} (a) compares the learning curves of Dynamic RL and conventional RL.
The vertical axis shows the average number of steps needed to reach the final goal,
calculated every 500 episodes.
The plot lines and shaded areas in the figure represent the mean and standard deviation
across all successful simulation runs.
When the agent could not reach the final goal within 200 steps, it was recorded as 200 steps.
It was observed that in all cases, the number of steps decreased as learning progressed.
After the environment changed just after the 20,000th episode, the number of steps increased once
but then decreased again.
The large standard deviation in conventional RL occurs because,
depending on the simulation run, the number of steps did not decrease from around 200 for some episodes.
The learning curve for (B) conventional RL without applying SAL was similar to that for (B-1) with SAL,
but slightly worse overall, although no figure is presented here.

Thus, Dynamic RL functions effectively even in a task where the agent must learn
to find necessary information, memorize it, and reflect it in its motions for appropriate future behaviors.
Dynamic RL does not require backward computation through time.
The actual computational cost for learning the actor network, compared to conventional RL (BPTT+SAL),
was 1:8.0 when using the author's Python and NumPy program without using a framework or a GPU. 
Nevertheless, the curve before the environmental change was similar to that of conventional RL.
Moreover, learning after the environmental change was significantly faster in Dynamic RL than in conventional RL.
In Dynamic RL, learning was faster after the environmental change compared to the beginning of the run;
whereas in conventional RL, it was slower.

Figure \ref{fig:LearningCurve_SW} (b) illustrates the learning curves for three cases in Dynamic RL:
(A) regular case, (A-1) without SRL for hidden-to-hidden connections, and (A-2) without SAL,
to evaluate the effects of SRL and SAL.
Note that even in the case of (A-1), SRL was still applied to the other connection weights.
%(A) proposed Dynamic RL, (A-1) Dynamic RL (SRL was not applied to the weights between hidden layers),
%(A-2) Dynamic RL (SAL was not applied).
Case (A) outperforms both (A-1) and (A-2), leading to the conclusion that both SRL and SAL are beneficial.
SAL seems to be generally more effective for fast learning, while application of SRL
to the connection weights between hidden neurons is particularly effective when the environment changes.

\begin{figure}[ht]
\centerline{\includegraphics[scale=0.7]{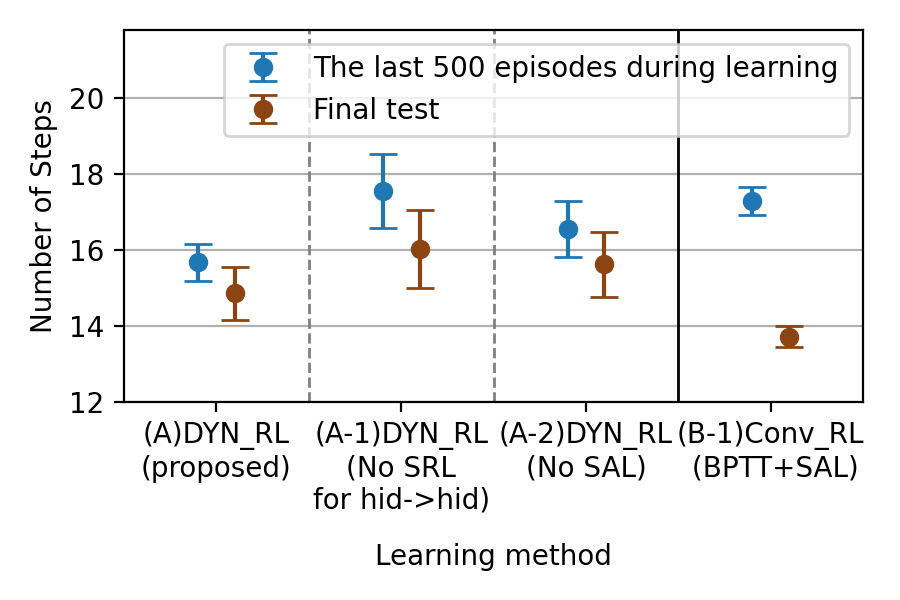}}
\caption{Comparison of the average number of steps in the final stage of learning and in the test
among four learning conditions.
The plots and the error bars indicate the mean and standard deviation over successful simulation runs.}
\label{fig:CompPerf}
\end{figure}
Figure~\ref{fig:CompPerf} compares the average number of steps in the last 500 learning episodes
to that in the final test, which consists of 46 episodes after learning.
This comparison is made across the four learning conditions presented in Fig. \ref{fig:LearningCurve_SW}.
% Figure~\ref{fig:CompPerf} shows its mean and standard deviation over the simulations where learning was successful.
In all cases, the number of steps in the test is lower than during learning.
This difference is due to the frequent starting positions at the periphery during learning.
In (B-1) conventional RL with SAL, the number of steps in the test is significantly lower than during learning
because no exploration noise was introduced in the test.

In case (A) Dynamic RL,
the agent's behavior became less exploratory through learning,
leading to a smaller average number of steps during the final stage of learning compared to (B-1) conventional RL.
However, the learning process in Dynamic RL was somewhat unstable,
sometimes causing sudden and significant increases in the number of steps,
which in turn increased both the average and standard deviation to that extent.
In the test, exploration was still present, and the average is larger than in conventional RL.
%exploration does not disappear, although the behavior becomes less exploratory through learning.
% but becomes smaller autonomously as the progress of learning as will be shown in Fig. \ref{fig:Task1_Loci}.
% Looking at this from the standpoint of Dynamic RL,
%This may have helped the Dynamic RL to adapt quickly to the new environment
%as shown in Fig. \ref{fig:LearningCurve_SW}.
%Furthermore, as previously mentioned, because the learning process is more unstable in Dynamic RL,
%the number of steps sometimes increases largely, and
%the standard deviation is larger in Dynamic RL than in conventional RL.
%As a result, in Dynamic RL, the average number of steps is smaller than in conventional RL in the final stage of learning
%but larger during the test.
%From the comparison of (A) to (A-1) or (A-2), we can see the usefulness of SRL and SAL also in the final stage of learning.
Additionally, in Dynamic RL, the absence of SAL or SRL in (A-1) or (A-2) resulted in an increase in the average and
standard deviation.

\begin{figure}[p]
\centering
%  \begin{minipage}[t]{0.4\linewidth}
  \begin{subfigure}[t]{0.45\linewidth}
  \centerline{\includegraphics[scale=0.3]{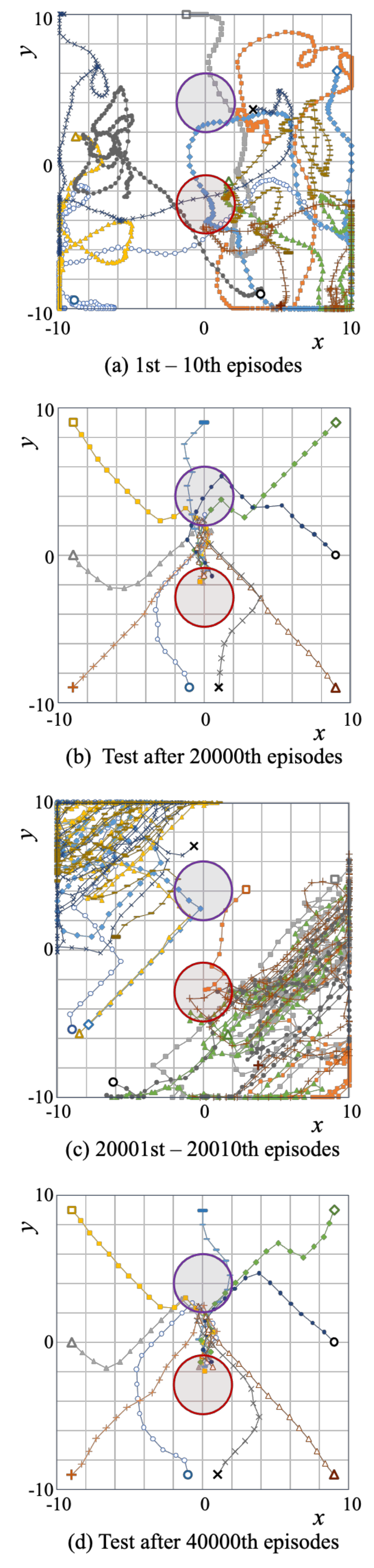}}
  \subcaption*{(A) DYN\_RL (proposed)}
  \end{subfigure}
 %\end{figure}
 %\begin{figure}[t]
%  \begin{minipage}[t]{0.4\linewidth}
  \begin{subfigure}[t]{0.45\linewidth}
  \centerline{\includegraphics[scale=0.3]{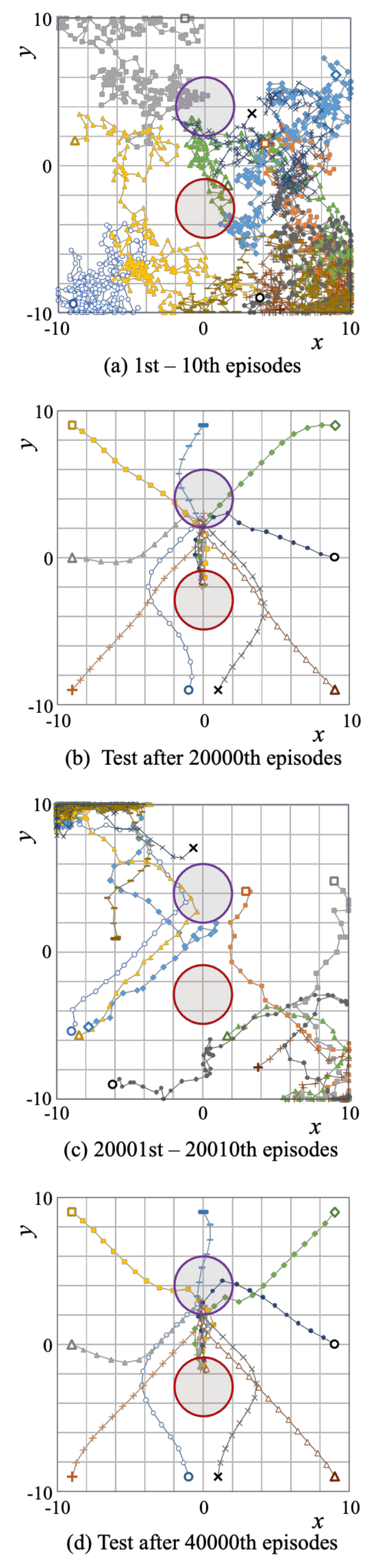}}
  \subcaption*{(B-1) Conv\_RL (BPTT+SAL)}
  \end{subfigure}
\caption{Comparison of sample agent behaviors at four stages in learning between (A) Dynamic RL and (B-1) conventional RL (with BPTT and SAL).}
\label{fig:Task1_Loci}
\end{figure}
%\newpage
Figure~\ref{fig:Task1_Loci} shows sample trajectories from a simulation run
that successfully learned for both (A) Dynamic RL and (B-1) conventional RL+SAL cases.
Since the number of steps varies largely among simulation runs in Dynamic RL, as shown in Fig.~\ref{fig:CompPerf},
the results for an average run are presented,
where the number of steps ranks 24th in the first test and 11th in the second test
out of 36 successful simulation runs for (A) Dynamic RL.
In (A) and (B-1), the cases for the same random sequences are shown.
The initial weights and biases in the RNNs, as well as the agent's starting position for each episode, were the same,
except for the weight matrix of the output layer and the spectral radius of the upper hidden layer's
self-feedback connection weight matrix in the actor network.
 
Figures~\ref{fig:Task1_Loci}(A)(a) and \ref{fig:Task1_Loci}(B-1)(a) show the agent paths in the first ten episodes.
The agent explored the entire field in both cases, but the exploration appeared different.
In (A) Dynamic RL, the paths were smoother due to the chaotic dynamics generated internally
with the large time constant in the upper hidden layer.
In (B-1) conventional RL with BPTT+SAL, they were jagged and irregular due to the random noise
introduced for exploration at each step.

Figures~\ref{fig:Task1_Loci}(A)(b) and \ref{fig:Task1_Loci}(B-1)(b) show the results of the simplified test
with nine initial locations
after 20,000 episodes of learning, just before the environmental change.
In this test, random exploration was stopped even in the conventional RL case.
Starting from any initial location, the agent first reaches the subgoal and then the final goal in both cases.
Dynamic RL could learn memory-required behaviors without using backward computation, such as BPTT.
However, the agent's paths in (A) were not as smooth as in the case of (B-1) conventional RL.
%As mentioned, i
In (A) Dynamic RL, the actor outputs were easily saturated and
the agent was apt to move diagonally although regularization was applied to avoid it. 
This problem needs to be solved.

Figures~\ref{fig:Task1_Loci}(A)(c) and \ref{fig:Task1_Loci}(B-1)(c) show the actual agent paths
for ten episodes just after the environmental change,
in which two actor outputs were swapped.
For instance, if the agent is located in the lower right direction of the subgoal
and intends to move toward the upper left direction, it actually moves to the lower right, which is opposite to its intention.
Consequently, the agent tends to gravitate toward the bottom right or top left corner.
Therefore, in (B-1) conventional RL with BPTT+SAL,
the agent's final position was in the area defined by $|x|, |y| > 9.0$ in 8 out of 10 episodes.
However, in (A) Dynamic RL, the agent's paths appear more exploratory,
and its final position was never in this area across the 10 episodes.
When the agent was not allowed to learn after the environmental change,
it moved to either the bottom right or top left corner from the starting position and remained there,
although no figure is presented here.
This suggests that Dynamic RL promotes exploration.
It can be inferred that the difference in exploration between the two RLs leads to variations in learning speed
after the environmental change.
Figures~\ref{fig:Task1_Loci}(A)(d) and \ref{fig:Task1_Loci}(B-1)(d) show the final simplified test results
after 40,000 episodes, i.e., after all learning was completed.
As in the case of (b), the agent regained the ability to reach the final goal in both cases,
and the paths were smoother, resulting in fewer steps in conventional RL.

Finally, the changes in dynamic characteristics during learning are compared.
Here, the exploration exponent is introduced, which is a type of Lyapunov exponent for the entire system, including
both the RNN and the physical system (See \ref{App:Expl_factor} for its calculation).
During learning, every 100 episodes, the exponent was computed for each of the nine starting points,
which are the same as in Fig.\ref{fig:Task1_Loci} (b) or (d), and the nine exponents were averaged.
It is calculated for each of the two environments that appeared before and after the environmental change,
referred to as Env1 and Env2, respectively.
%Env1 and Env2 are the learning environments before and after the environmental change respectively.

\begin{figure}[tb]
\centerline{\includegraphics[scale=0.6]{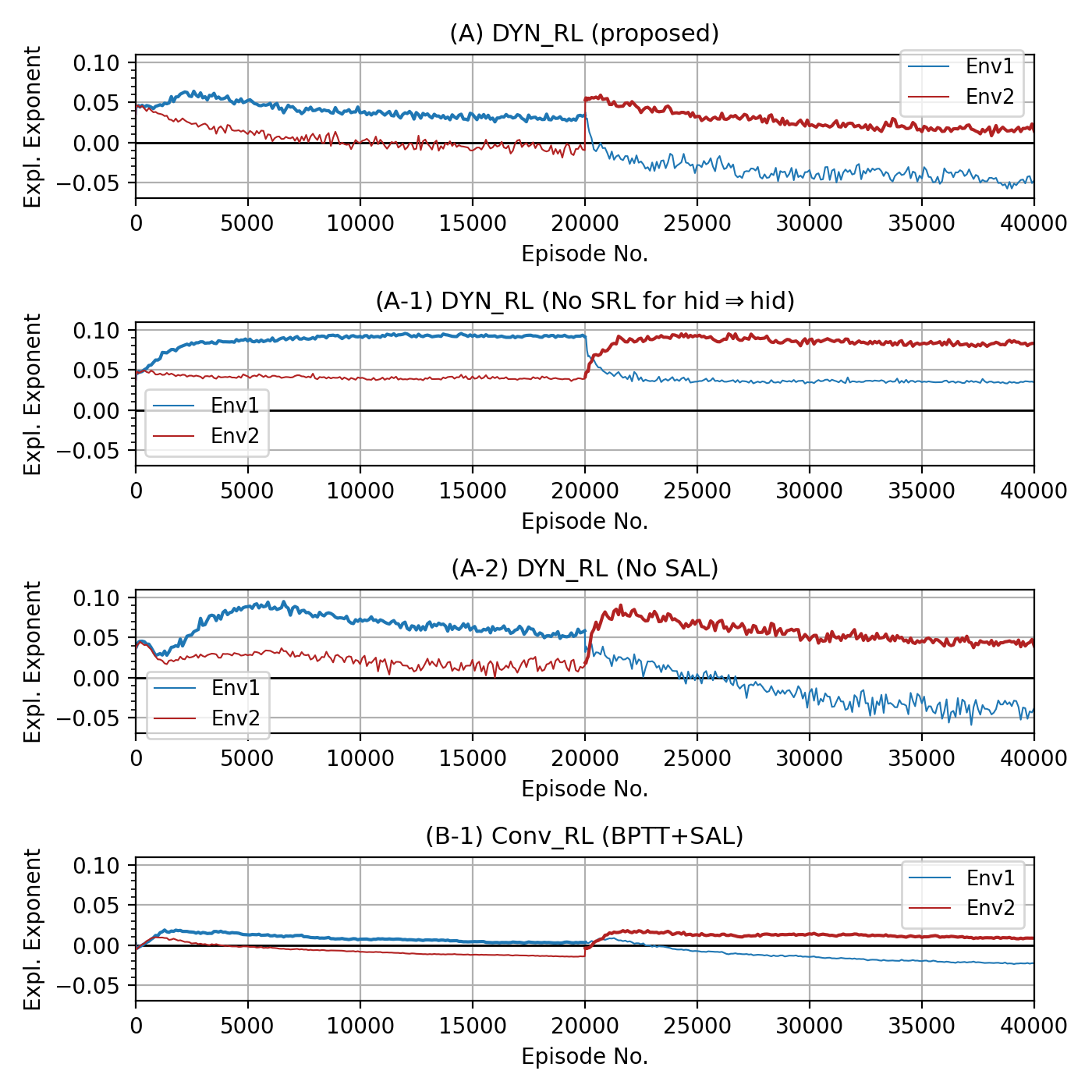}}
\caption{Comparison of changes in exploration exponents in the sequential navigation task
among four learning conditions.
Refer to the text for the definition of the exploration exponent.
Env1 and Env2 are the learning environments before and after the environmental change, respectively.
Thick lines indicate an exponent in the learning environment: Env1 before environmental change and
Env2 after that.}
\label{fig:ExplExponent}
\end{figure}

Although the exponent fluctuated significantly, the average over successful simulation runs
reveals the characteristics of each learning condition, as shown in Fig.~\ref{fig:ExplExponent}.
The four graphs show how the exploration exponent changes as learning progresses
under the four learning conditions.
In (A), (A-1), and (A-2), the starting value is larger than in (B-1)
due to the differences in the output connection weights
and the spectral radius of the self-feedback connection weights in the upper hidden layer.
%That is because the output connection weights in the actor network are not zero,
%and furthermore, the spectral radius of the initial self-feedback connection weights in the higher hidden layer
%was larger in (A), (A-1), and (A-2).
%The way of value change differs among the four learning conditions but
%In all cases, the value in the Env1 is larger than in the Env2 in the first half of the learning,
%and after the environmental change,  the relationship between the larger and smaller values is reversed.

In the proposed Dynamic RL (A), the exponent in the learning environment first increased
and then gradually decreased while maintaining the value in the learning environment
(Env1 for the first half of the learning episodes and Env2 for the second half)
greater than that in the non-learning environment.
When SRL was not applied to the hidden-to-hidden connections in (A-1),
the exponent did not decrease as learning progressed.
This suggests that SRL facilitates convergence in the dynamics by reducing each neuron's sensitivity
around the state transitions when the TD-error is positive.
Conversely, when SAL was not applied in (A-2), the exponent initially decreased.
This indicates that SAL prevents the system dynamics around the actual state transitions
from converging excessively by increasing each neuron's sensitivity.

Moreover, after the environment changed, the exponent for the learning environment Env2 in (A)
increased much faster than in the other cases.
%In many cases, the agent often went to the corner of the field in the non-learning environment,
%and the locations on the two agent paths for the computation of the exponent overlapped exactly.
This aligns with the rapid adaptation in the learning curve observed in Fig.~\ref{fig:LearningCurve_SW}.
The slow adaptation observed in (A-1) and (A-2) suggests that the quick increase of the exponent
by both SRL and SAL was the key to quick learning after the environmental change.
While the exponent for the new learning environment, Env2, increased,
the exponent for Env1, which was newly transitioned into a non-learning environment, decreased.
This indicates that the agent enhanced its exploratory behavior, specializing in the learning environment.
%In (A-1), the exponent changed at first in each environment but did not change so much after that.
%This would be because SAL first increased the sensitivity of each hidden neuron at first in the learning environment,
%and then it was not applied when the sensitivity became large enough.
%This suggests that SRL decreased the sensitivity according to the progress of learning.
%In the case of (A-2) in which SAL was not applied,
%the exponent became small once in the early phase of learning, and then gradually became large.
%This and  \citep{Sensitivity} suggest that SAL avoided low sensitivities in hidden neurons
%and accelerated learning as shown in Fig. \ref{fig:LearningCurve_SW}.
%Since the fluctuation is smaller in (A-1) than in (A), it is possible that SRL causes the learning unstable.
%As for the large fluctuation in (A-2), the possibility that the small number in successful simulation runs
%can cause it should be also considered.

Meanwhile, in (B-1) BPTT with SAL, the trend in the exponent change was similar to that of (A),
but the exponent started from a smaller value and stayed close to zero.
This low exponent in the learning environment may have led to stable learning but slow adaptation
after the environmental change.
%This would be because since the conventional RL committed exploration to the external random noise,
%the initial connection weights were set so that the initial exponent was small,
%and learning did not directly control the dynamics.
%The exponent at the end of learning in each learning environment was larger in Dynamic RL than in conventional RL.
%This may cause the instability of learning in Dynamics RL.
%The change of the exponent after the environmental change was not so fast.

\subsection{Slider-Crank control (Dynamic pattern generation) task}
\begin{figure}[htb]
\centerline{\includegraphics[scale=0.43]{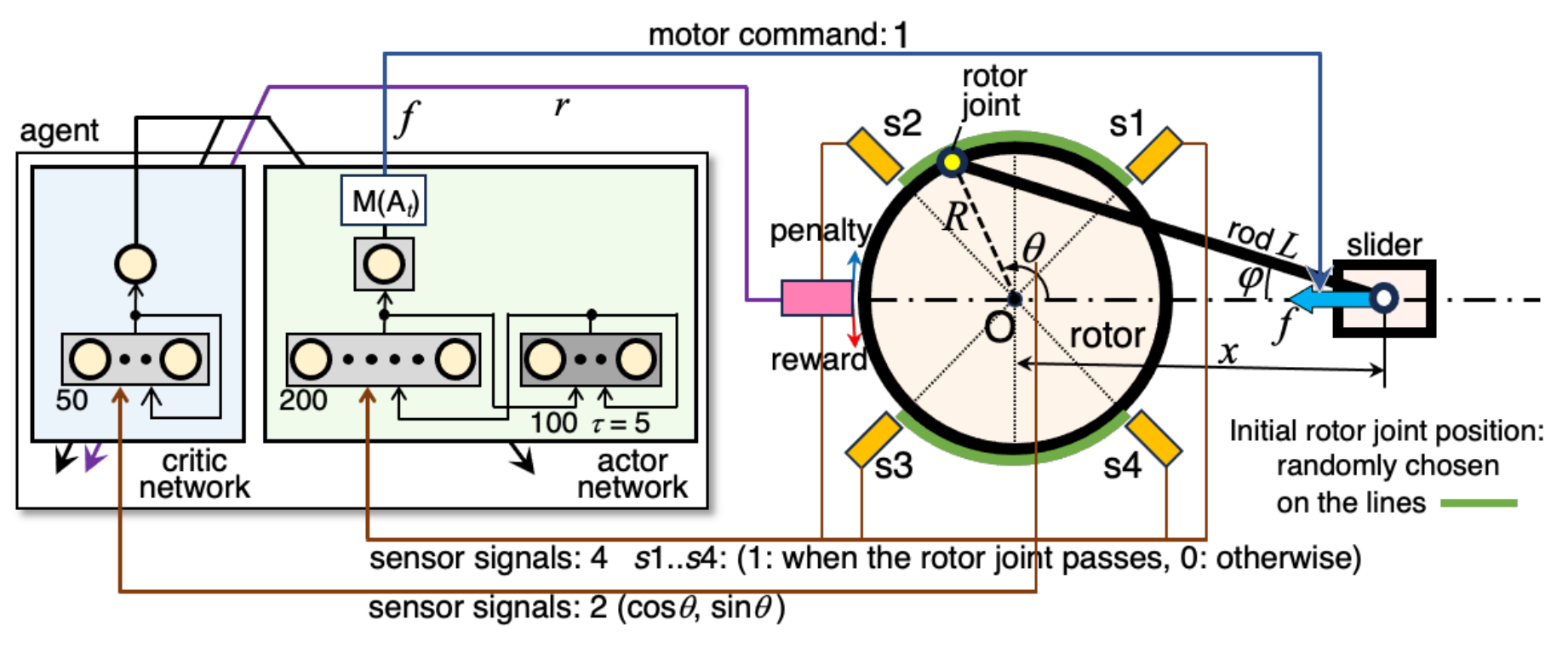}}
\caption{Slider-crank control task as a dynamic pattern generation task and the RNN used.
The agent applies a force $f$ to the slider in the direction of the thick blue arrow
(which can be positive or negative), causing the rotor to rotate.
There are four joint-detecting sensors (s1,...s4), and only the four binary signals from them
are the input to the actor network.
The initial rotor joint angle $\theta_0$ is randomly chosen within the ranges indicated by the two green arcs at the beginning of each episode.
}
\label{fig:Crank}
\end{figure}

To evaluate whether the proposed Dynamic RL works in a task where a learning agent
is required to generate dynamic patterns,
a slider-crank control task, as shown in Fig.~\ref{fig:Crank}, was employed as the next learning task.
In this task, an agent learns to apply an appropriate time-varying force $f$ in the $x$-direction to the slider.
This force is then transmitted to the rotor through the connecting rod,
causing the rotor to rotate around its central axis O.

The actor network receives only four binary sensor signals: $s_1$ to $s_4$.
Each signal is 1.0 only when the rotor angle $\theta$ passes through a specific angle: $\pi/4, 3\pi/4, -\pi/4$, or $-3\pi/4 $ radian
respectively, regardless of the rotational direction.
The actor network receives the four signals and outputs only one motor command
representing the force $f$ applied to the slider in the $x$-direction.
Therefore, the agent cannot always perceive the rotational angle $\theta$ or the slider's location $x$,
nor can it directly perceive the rotational direction or speed.
The agent has to generate a force sequence
that is not defined as a static function of its four inputs.
The actor output is modified by multiplying it by 1.25 and clipping it between -1.0 and 1.0
to convert it into a motor command (force $f$).
This is expected to prevent the output from entering the saturation range of the neuron's activation function.
The uniform disk-type rotor, with mass, makes this system dynamic
while considering the negligible masses of the slider and link.
The equations of motion for this system are provided in \ref{App:Eq_Crank},

In each episode, the initial rotation angle $\theta_0$ is set randomly
to an angle from $-3\pi/4 <= \theta_0 <= -\pi/4$ or $\pi/4<=\theta_0<=3\pi/4$ radian.
The initial rotational speed is 0.0.
Since the agent does not perceive the joint location directly,
it cannot determine the current rotational direction or identify the necessary direction of force
to rotate the rotor in the rewarded direction until it receives the first ``on'' signal.
%the initial rotor angle is $-3\pi/4 <= \theta <= -\pi/4$ and
Since Dynamic RL was not applied to the critic network,
it was designed to receive $cos \theta$ and $sin \theta$ as sensor inputs,
which are sufficient for identifying the rotor angle, for simplicity.
When the angle passes through $\pi$ with a positive rotational speed, the agent receives a reward of 0.15,
and when with a negative rotational speed, it receives a penalty of -0.15.
In each episode, the agent performed 400 steps of force loading,
and the episode did not end even though the agent received a reward or penalty halfway through.
In one simulation run, 10,000 episodes of learning were conducted.
After 5,000 episodes, the reward sign was reversed as an environmental change.
Tests were performed after 5,000 and 10,000 episodes.
In each test, the rotor started from each of 50 initial angles at intervals of $0.02\pi$,
and the average revolutions per episode were calculated.
Learning is defined as a ``failure'' if the average is less than 20 in the first test or more than -20 in the final test;
otherwise, it is defined as a ``success.''
As well as the previous task, 40 simulation runs were performed for each learning condition.
The parameters used here were almost the same as in the previous task,
but some of them are different as in Table \ref{Table:Parameters2}.\vspace{5mm}

\begin{table}[ht]
  \begin{center}
    \caption{Parameters used in the dynamic pattern generation (slider-crank control) task,
    excluding the same ones as in the previous task.
    %that differ from the previous task
    For the excluded ones, refer to Table \ref{Table:Parameters1}.}
    \vspace{2mm}\small
      \begin{tabular}{c|c} \toprule
	Target Sensitivity for SAL  $s_{th}$ in Fig.~\ref{fig:DynamicRL}& 1.6\\ \midrule
	Rate of Regulation $\eta_{reg}$ in Eq.~(\ref{Eq:Regularize}) & $1e^{-7}$\\ \midrule
	Rate of Raising Critic $\eta_{raise}$ in Eq.~(\ref{Eq:Raise_Critic}) & 0.0002\\ \midrule
	Truncated Number of Steps in BPTT & 10\\ \midrule
      \end{tabular}
    \label{Table:Parameters2}
  \end{center}
\end{table}

\begin{figure}[ht]
\centerline{\includegraphics[scale=0.7]{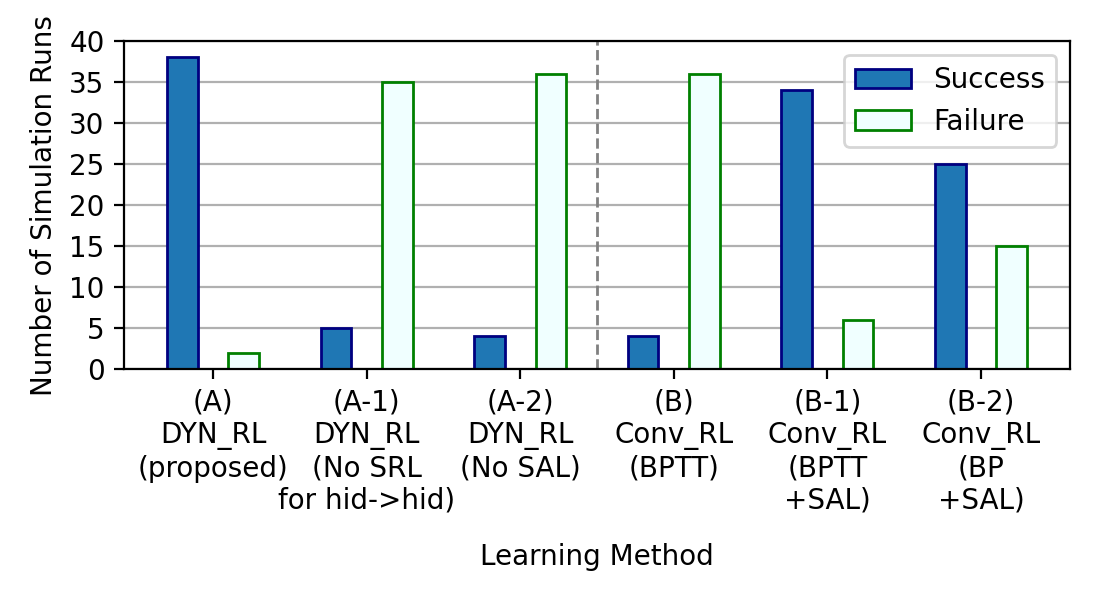}}
\caption{Comparison of success rate among six learning conditions in Table~\ref{Table:CompSet} in the slider-crank control task.}
\label{fig:CompPerf1_CR}
\end{figure}
Figure \ref{fig:CompPerf1_CR} shows the counts of ``success'' and ``failure'' runs for each learning condition.
As shown, (A) Dynamic RL achieved the highest success rate, while (B-1) conventional RL (BPTT + SAL) had
the second-highest rate.
In contrast, (B-2), which did not use backward error propagation through time, still achieved a high success rate.
In the other cases, including the case (B) where SAL was not applied in conventional RL,
the success rate was notably low.
This indicates that both SRL and SAL are essential in Dynamic RL, and 
a method to ensure sensitivity, like SAL, is also necessary in conventional RL.

\begin{figure}[t]
\centerline{\includegraphics[scale=0.6]{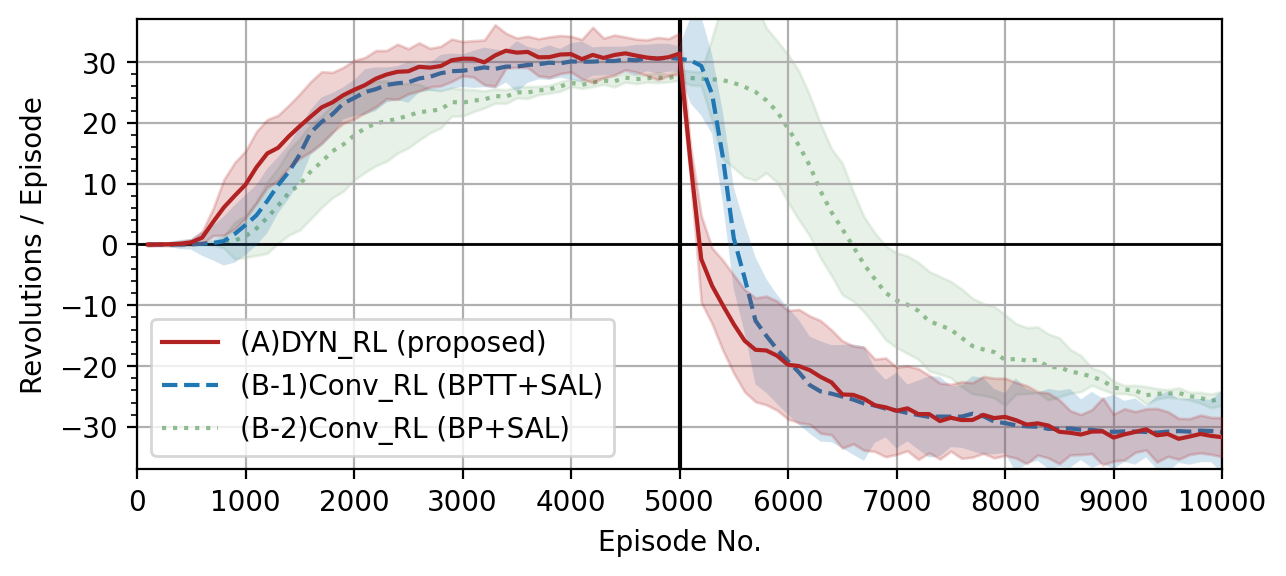}}
\caption{Comparison of changes in the average number of revolutions per episode, calculated over every 100 episodes, 
as learning progresses among three learning conditions in the slider-crank control task.
Each line and shaded area indicate the mean and standard deviation of the number over successful simulation runs, respectively.}
\label{fig:LearningCurve_Crank}
\end{figure}

\begin{figure}[pth]
\center
\centerline{\includegraphics[scale=0.65]{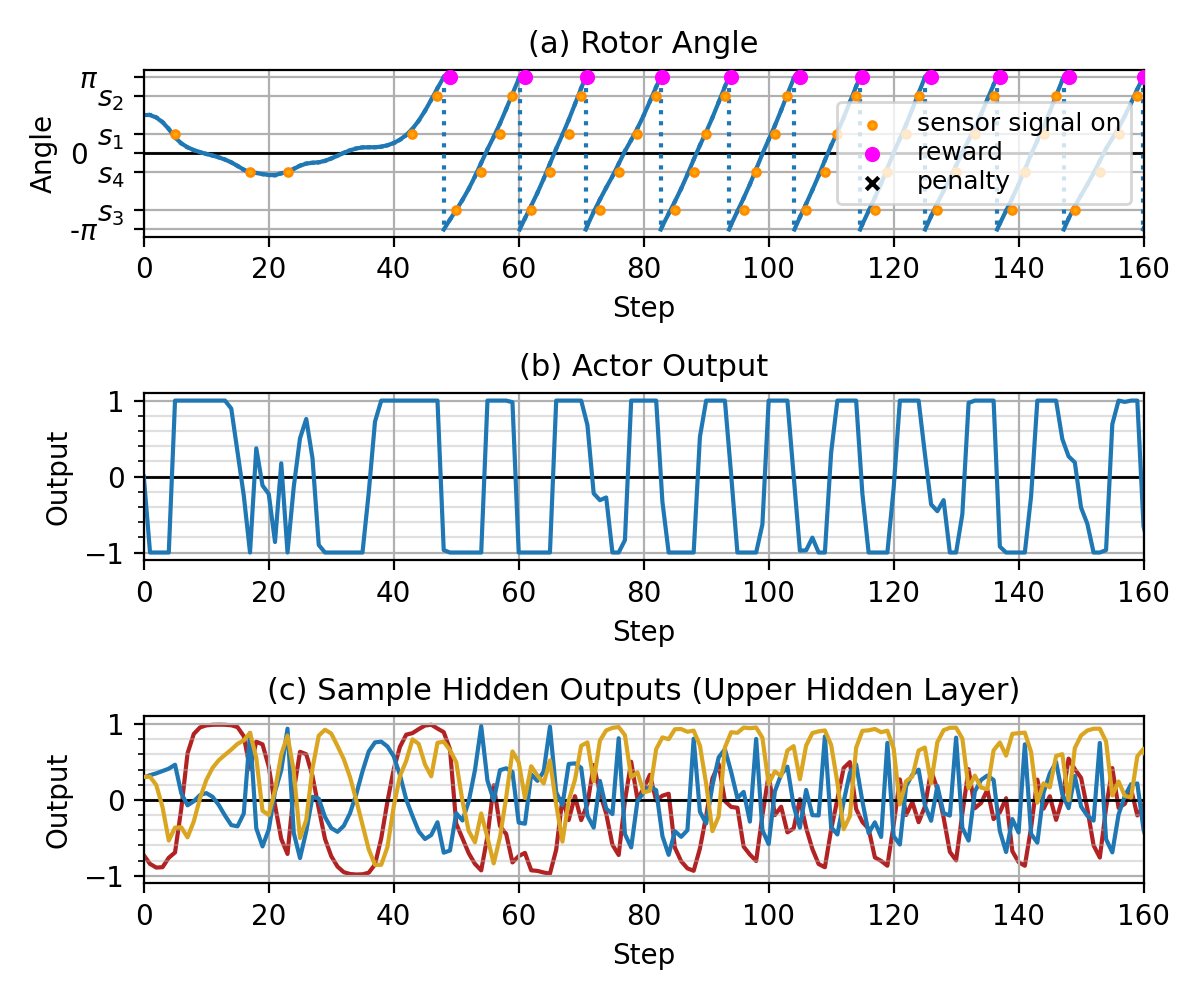}}
\footnotesize{(A) DYN\_RL}
\vspace{3mm}
\centerline{\includegraphics[scale=0.65]{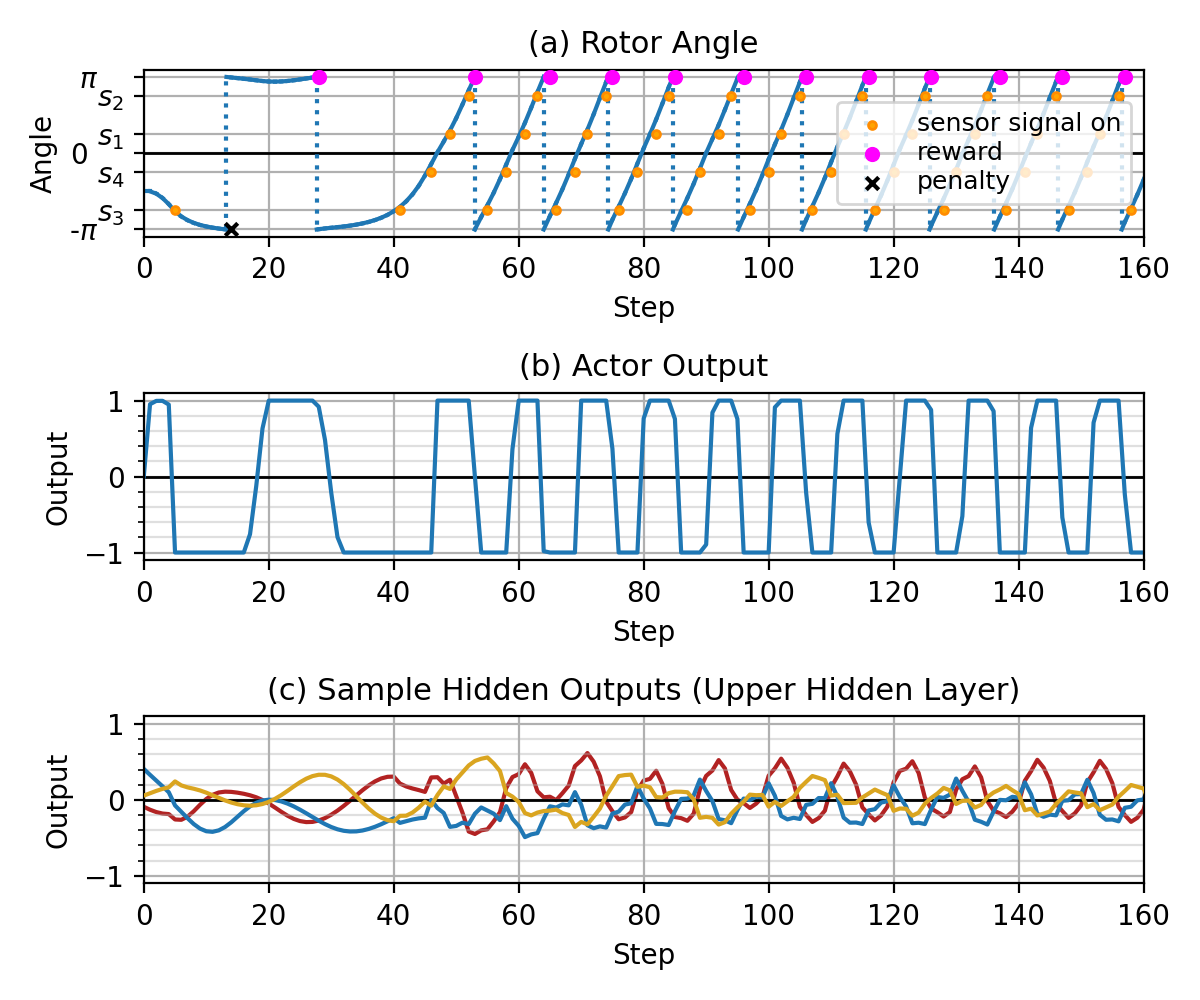}}
\footnotesize{(B-1) Conv\_RL (BPTT+SAL)}
\caption{Comparison of changes in (a) rotor angle with four sensor signals and a reinforcement signal
(reward or penalty), (b) actor output, and
(c) outputs of three sample upper hidden neurons during a test episode after 5,000 episodes of learning
between (A) Dynamic RL and (B-1) conventional RL (with BPTT and SAL) in the slider-crank control task.
The initial rotor angle was different between (A) and (B-1) because these show the cases
where the agent first rotated the rotor in the opposite direction to the rewarded one.}
%and then changed it after receiving the first `on' signal.}
\label{fig:Epi_CR_before}
\end{figure}

\begin{figure}[pth]
\center
\centerline{\includegraphics[scale=0.65]{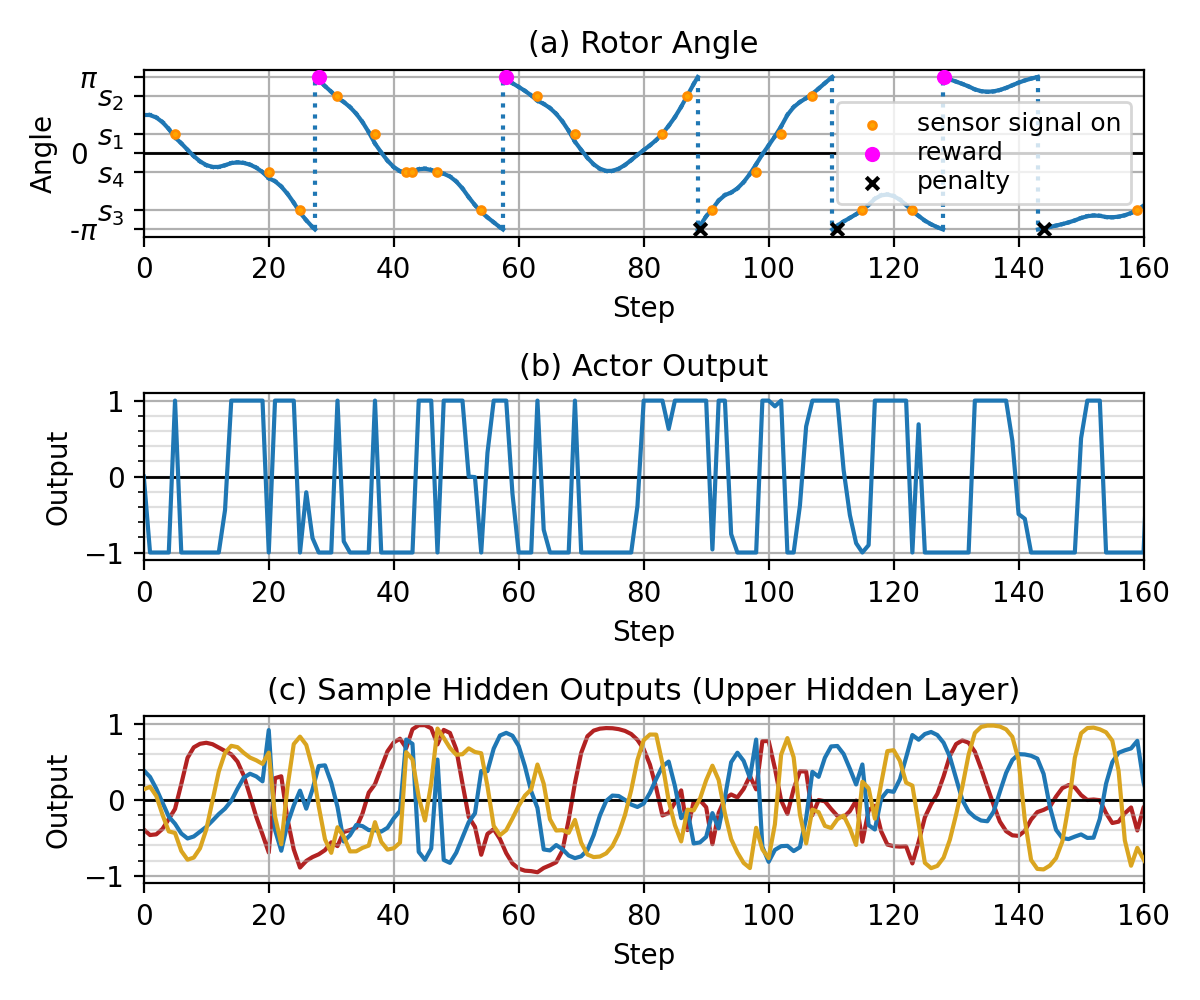}}
\footnotesize{(A-I) DYN\_RL (after 5100 episodes)}
\vspace{5mm}
%\caption{Changes in the angle and some outputs of neurons in the 5100th episode in dynamic RL.}
\centerline{\includegraphics[scale=0.65]{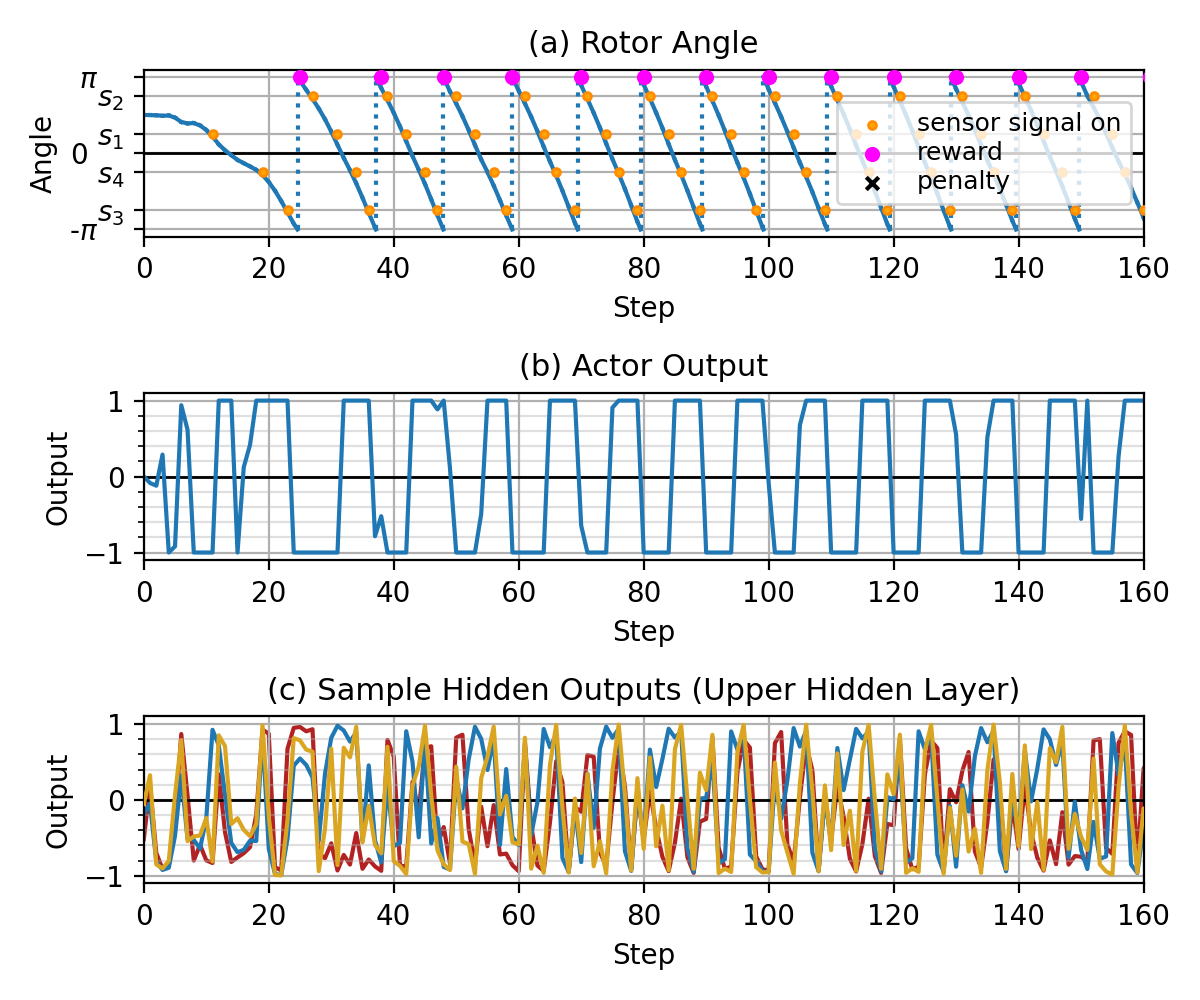}}
\footnotesize{(A-II) DYN\_RL after learning}
\caption{Changes in (a) rotor angle with four sensor signals and a reinforcement signal (reward or penalty),
(b) actor output, and (c) outputs of three sample upper hidden neurons
at two phases after the environmental change in the cases of (A) Dynamic RL.
(A-I) represents the state soon after the environmental change, and (A-II) reflects the state after all learning had finished.}
\label{fig:Epi_CR_after}
\end{figure}

Figure \ref{fig:LearningCurve_Crank} shows the change in the number of revolutions per episode
% in the direction of reward per episode
for the three cases with a high success rate, as shown in Fig.~\ref{fig:CompPerf1_CR}.
When the proposed Dynamic RL (DYN\_RL) was applied,
the rotational speed gradually increased, and the average number of revolutions per episode reached approximately 30.
Learning progressed slightly faster compared to the case of conventional RL using BPTT,
despite not needing backward computation through time.
The ratio of the actual computational cost for learning the actor network,
compared to conventional RL (BPTT+SAL), was 1:4.9.
The difference from the previous task is mainly
due to the difference in the truncated time steps in BPTT: 10 in the sequential navigation task and 5 in this task.

Shortly after the rotational direction for the reward was reversed as the environmental change after 5,000 episodes,
% if the agent behaved the same as before the environmental change, the sign was reversed
% while maintaining the absolute value is the same.
the number of revolutions decreased, and the average over the 5,100-5,200 episodes was already negative.
As in the previous task, the learning speed immediately after the environmental change was considerably faster
than in the case of conventional RL.
In conventional RL with only BP, the success rate was reasonable,
but the learning speed and performance after learning were much lower compared to the case with BPTT.

Figure \ref{fig:Epi_CR_before} illustrates the changes in several states during a sample test episode just before the environmental change
for both cases: (A) Dynamic RL and (B-1) conventional RL using BPTT and SAL.
Specifically, the states include (a) rotor angle with sensor signals and reinforcement signal, (b) actor output,
and (c) three sample neuron outputs from the upper hidden layer.
%In this task, the agent could not know the direction of rotation until either sensor signal became on.
The figure depicts the case in which the initial rotational speed caused by the agent's force load was negative,
i.e., opposite to the rewarded direction, for one of the two initial angles: $-\pi/2$ or $\pi/2$.
Therefore, the agent had to change its direction after detecting an ``on'' sensor signal.
From (a), it can be observed that the agent changed the rotational direction in both cases
and then maintained it thereafter.
After the change, the outputs of the actor and hidden neurons changed almost periodically, as shown in (b) and (c).
In the case of Dynamic RL, the outputs changed more irregularly than in conventional RL.
%that would be because in the Dynamic RL, the outputs included irregular factors due to the chaotic dynamics.
%On the other hand, since this was a test phase, irregular factors seem smaller in the case of conventional RL, 
This observation was made during the test phase; however during learning,
external random noise for exploration was added
to the actor outputs in the case of conventional RL.
Another difference is that the hidden neuron outputs in Dynamic RL have a larger variation range,
though the underlying reasons or influences have yet to be investigated.

Figure~\ref{fig:Epi_CR_after} shows the changes in states in two phases after the environmental change
in the case of Dynamic RL.
Figure~\ref{fig:Epi_CR_after}(A-I) shows the state just after 5,100 episodes, shortly after the environmental change,
while Fig.~\ref{fig:Epi_CR_after}(A-II) depicts the state in the test phase after all learning was completed.
In both cases, the initial rotor angle was the same as in Fig.~\ref{fig:Epi_CR_before}(A).
In Fig.~\ref{fig:Epi_CR_after}(A-I), the actor output and hidden outputs changed irregularly, causing the rotor to rotate
both clockwise and counter-clockwise irregularly.
In contrast, in Fig.~\ref{fig:Epi_CR_after}(A-II) where learning had progressed, the rotor rotated in the opposite direction
compared to Fig.~\ref{fig:Epi_CR_before}(A), and the actor output, along with the hidden outputs, changed
almost periodically sometime after the start of the episode.
However, the periodic pattern of the upper hidden neurons in Fig.~\ref{fig:Epi_CR_after}(A-II) differs
from that in Fig.~\ref{fig:Epi_CR_before}(A),
%This means that the fast learning after the environmental change in Dynamic RL
%was not because the internal periodic pattern did not change.
%revealing no common features in the internal periodic patterns
%before and after the environmental change.
making it hard to explain the rapid adaptation after the environmental change in Dynamic RL
based on the retention of internal periodic patterns.

\section{Discussion and conclusion}\label{Sec:Discussion}
In this paper, the author proposed Dynamic Reinforcement Learning (Dynamic RL), a new RL framework,
expecting it to be a fundamental technique for the emergence of `thinking.'
This paper could not clearly show its critical advantages in this regard. 
%This RL differs significantly from conventional RL in terms of exploration,
%which is performed as chaotic dynamics produced through the process of generating continuous motor commands
%rather than stochastic action selection separately from action generation.
%as the first step towards its emergence,
However, the author demonstrated that despite the major shift in learning from static to dynamic,
this unconventional RL approach effectively functions as RL in two simple dynamic tasks.
Learning recurrent connection weights among hidden neurons without error backpropagation
improved learning performance, which was not achievable with previous chaos-based RL \citep{IJCNN2015}.
The autonomous change in the state transitions from ``more irregular'' to ``more rational'' through learning observed in the simulations
%He has also shown that through the direct learning of the system dynamics that generate motions in which explorations are inherent,
%the proposed RL changes the dynamics autonomously from more irregular to more rational while maintaining chaotic dynamics.
appears to have solved the balance between exploration and exploitation \citep{Sutton1998}
and suggests the potential for the emergence of thinking.
The learning characteristics of Dynamic RL, as observed in the simulation results of the two tasks,
are summarized first, followed by a discussion of its possibilities and challenges from broader perspectives.
% towards the emergence of thinking are discussed.
Finally, the risks and concerns of Dynamic RL are discussed.

\subsection{Summary as one method of Reinforcement Learning}
Dynamic RL allows for significantly lower computational costs per step than conventional RL
due to the absence of backward computation for learning, particularly along the time axis.
Nevertheless, the success rate was almost the same as that of (A) and (B-1)
in both Figs.~\ref{fig:CompPerf1_SW} and \ref{fig:CompPerf1_CR},
and the learning curve with episodes as the horizontal axis was also comparable to that of conventional RL
using BPTT and SAL as (A) and (B-1)
in both Figs.~\ref{fig:LearningCurve_SW} and \ref{fig:LearningCurve_Crank}, regardless of the learning task.
Furthermore, as for adapting to new environments, conversely, Dynamic RL was considerably faster.
The agent's behaviors became more exploratory immediately after entering a new environment
and less exploratory as learning progressed thereafter, as shown in Fig.~\ref{fig:Task1_Loci}(A).
This was also reflected in the exploration exponent,
a type of Lyapunov exponent for the entire system, including the external world,
as shown in Fig.~\ref{fig:ExplExponent}(A) in the sequential navigation task.
In the slider-crank control task, it was reflected in the internal state change of the RNN,
as shown in Figs.~\ref{fig:Epi_CR_before} and \ref{fig:Epi_CR_after}.
%This could be because Dynamic RL could effectively use what it had learned before
%and/or the hidden neurons could learn directly without using back-propagated errors,
%but further studies are needed.

Comprehensively considering the results of varying the application of
SAL (Sensitivity Adjustment Learning) or SRL (Sensitivity-Controlled RL),
the author suggests that the following mechanisms are functioning.
If neurons are linear, larger weights generally increase sensitivity.
However, due to the saturation property of each non-linear neuron,
excessively large weights often reduce sensitivity conversely
by decreasing $f'(U)$ in Eq.~(\ref{Eq:sensitivity}).
This generally decreases the Lyapunov exponent of the system and also the learning activities.
Such problems occur when the initial weights are very large
or often when another learning method (SRL in this case) progresses,
as shown in the early phase of Fig.~\ref{fig:ExplExponent}(A-2).
SAL maintains the sensitivity of each neuron,
thereby preserving the exploration exponent positive (chaotic) in the learning environment,
as shown in Fig.\ref{fig:ExplExponent}(A).
This prevents excessive convergence of the system and a decline in learning activity.
Therefore, the absence of SAL leads to a decrease in both the learning success rate and learning speed, as shown in
the difference between (A) and (A-2) in Figs.~\ref{fig:CompPerf1_SW}, \ref{fig:LearningCurve_SW}, 
and \ref{fig:CompPerf1_CR}.
This is the case not only for Dynamic RL but also for BPTT, as shown in the difference between (B) and (B-1)
in Figs.~\ref{fig:CompPerf1_SW} and \ref{fig:CompPerf1_CR},
which is compatible with the results in the previous work \citep{Sensitivity}.

As learning progresses, while avoiding over-convergence by using SAL,
SRL adjusts the dynamics around the state transitions with positive TD-errors to be more convergent;
thus, the state transitions become more reproducible.
Conversely, it adjusts the dynamics around the state transitions with negative TD-errors to be more exploratory.
Since the irregularity sometimes increases and sometimes decreases depending on the TD-errors,
it seems that the irregularity remains unchanged in total.
However, actually, since the frequency of better state transitions increases,
%for state transitions where irregularity decreases, reproducibility increases,
%while for those where irregularity increases, reproducibility decreases.
the irregularity around actual state transitions decreases,
as the exploration exponent in Fig.~\ref{fig:ExplExponent}(A).
%This function of SRL clearly appears in the differences in the change of the exploration exponent between
%Figs.~\ref{fig:ExplExponent}(A) and (A-1).
Moreover, the state transitions that are adjusted to be more reproducible had positive TD-errors.
Therefore, SRL makes the state transitions not only ``less irregular'' but also ``more rational.''

%SAL always prevents over-convergence and maintains chaos in the current learning environment.
%That ensures that an agent continues to explore greater or less and remains flexible.
%Furthermore, in a new environment, since the TD-error is negative more often,
%SRL also makes the current system dynamics more chaotic.
After the environment changed during learning, the agent rapidly resumed exploration,
and its performance quickly recovered in Dynamic RL. 
In the navigation task, the dynamics immediately before entering the new environment were neither chaotic nor exploratory
for the new environment Env2,
as the exploration exponent for Env2 was slightly below zero at 20,000 episodes in Fig.~\ref{fig:ExplExponent}(A).
Observing the agent's behavior from two slightly separated initial positions, 
the agent moved to the same corner of the field and remained trapped there in most cases.
When the agent moved toward or was trapped in the corner, 
%its behavior was worse than expected, and
the TD-error was negative very frequently.
In the slider-crank control task, due to the reversal of the reward's sign, large negative TD-errors frequently occurred. 
Accordingly, in both cases, it can be argued that SRL restored the strong chaotic nature of the system dynamics
with the help of SAL and resumed exploratory behaviors,
%it can be understood that the rapidly restoring chaotic nature of the system dynamics
%and resuming exploration behaviors by SRL and SAL shortly after the environmental change,
resulting in the quick recovery of performance.

For agents, the dynamics are shaped not only by the RNN
but by the entire system integrated with the outside world, including the physical systems of the agents and their environment.
Nevertheless, the gap is interesting where only adjusting or controlling the sensitivity of individual neurons locally
by updating their weights and bias can globally control the system dynamics,
thereby regulating the system and enabling the agent to respond effectively to the external world.
In this paper, the sensitivity targets for SAL are set as appropriate values
at 1.3 for the navigation task and 1.6 for the slider-crank control task.
A value of 1.0 was appropriate for the sensitivity target in the preceding study
investigating SAL in supervised learning with BP or BPTT \citep{Sensitivity}.
In reward-modulated Hebbian learning using chaotic dynamics, 
performance was optimized near the ``edge of chaos'' \citep{NN_Matsuki}.
The author is considering the possibility that a large value may be suitable for RL's exploration, which requires stronger chaoticity.
%However, further research is needed on the sensitivity target,
%considering its relation to thinking that requires autonomous state transitions,
%including unexpected ones in inspiration or discovery.

For massively parallel and flexible systems, asynchronous pulse-and-analog coexisting systems would be far superior
to a synchronous digital system.
The introduction of Dynamic RL into brain models or brain-like hardware
based on such systems offers significant advantages.
Local learning without backward computation significantly reduces communication between neurons,
as well as the computational cost in each neuron, compared to conventional RL using BPTT.
In particular, eliminating backward computation through time is highly beneficial,
as it removes the need for time management and memory to store past states.
Furthermore, a separate random number generator for stochastic selection is now unnecessary.
A concern remains on how to build an RNN with connections that initially have large, random weights
to make the system dynamics chaotic.
The author expects SAL will enlarge small, irregular initial weights,
as in the case of supervised learning in \citep{Sensitivity}.
The introduction of chaotic neurons \citep{Aihara} would help it.
In this paper, Dynamic RL is built as a discrete-time model to compute on a digital computer,
but it would not be difficult to implement it on a continuous-time system.

However, even though we see Dynamic RL as regular RL, there are still some problems and concerns.
First, learning was more unstable than conventional RL, showing occasional performance drops during learning.
%and sometimes the learning performance dropped suddenly during learning.
This resulted in larger deviations in learning performance in Dynamic RL than in conventional RL,
as shown in Fig.~\ref{fig:CompPerf}.
%In Dynamic RL, weights and biases are updated to control the convergence or divergence of the flows
%around actual state transitions, but at the same time, by this update,
%not only the neighbors but also the transitions themselves are changed together even when TD-error is positive.
%This may cause the observed instability in learning.
%, and its influence needs to be investigated.
%As for these two problems, because the exploration is inherence in the action
%and also from the view of life-long learning, it could be considered unavoidable to some extent.
% However, if humans are learning similarly, there should be more room for improvement.

The second issue is that the actor outputs are often saturated, making it difficult to fine-tune them.
This can be seen, for example, in the agent's tendency to move in the 45-degree diagonal direction in Fig.~\ref{fig:Task1_Loci}(A)(b,d)
compared to the conventional RL case in Fig.~\ref{fig:Task1_Loci}(B)(b,d),
though they were improved to some extent by regulation learning as in Eq.~(\ref{Eq:Regularize}).
%A more fundamental solution is needed.

The third issue is that, in Dynamic RL, the exploration factor cannot be separated from the actor output.
Therefore, as shown in Fig.~\ref{fig:CompPerf}, the performance in the test is worse than in conventional RL,
where the actor outputs can be used without adding exploration noises.
 %Some solutions could narrow this gap, which would be, however, somewhat inherently inevitable.
The fourth is that there are many parameters for learning, and adjusting them is not easy.
In this sense, the author's computational power (iMac 2020, Intel Core i9 3.6GHz) did not allow systematic
and sufficient optimization in this study.

The author currently has several major challenges in mind, along with ideas for solving some of them,
regarding this research.
However, due to the risk discussed in subsection \ref{subsec:Risk},
he would rather refrain from writing about that.
%The main remaining challenges are as follows.
%Herein, the critic was learned by the conventional RL using BPTT.
%However, we will be able to say that Dynamics RL has been established
%only when critic learning is performed within the framework of Dynamic RL.
%To achieve this, further study is required,
%including the introduction of a novel architecture like TD3 \citep{TD3, TD3_Matsuki}.
%Another challenge is the acquisition of multistep state transitions through RL.
%It was demonstrated that Dynamic RL can learn past processes without going back through time,
%but there must be a limitation to that.
%This would need an additional technique; local memories, such as causality traces \citep{CausalityTraces},
%would be the key to solving it.
%%The network architecture should also be considered, including such as TD3 \citep{TD3,TD3_Matsuki}.
%The Introduction states that Transformers surpass traditional AI.
%They significantly differ from the proposed Dynamic RL and conventional RL.
%%but demonstrates that it has much to do with `thinking.'
%However, Transformers are essential for achieving `thinking' due to their capabilities in generative AI. 
%Further exploration of the two and also their integration is needed.

\subsection{Towards the future from a higher perspective}\label{subsec:Future}
Even from the broader perspective of neural network (NN) learning not limited to RL,
the Dynamic Learning approach used in Dynamic RL is still quite different in quality from conventional approaches.
The state vector of an NN represents a point in its internal state space.
In general NN learning, a gradient-based method or a derivative of it determines
how to move this point in space based on some value or cost function, such as an error function,
whose inputs are the NN outputs.
Then, how to change the weights and biases is determined.
Learning considers only the current point and usually does not consider its temporal changes.
BPTT applies the same method to past states or points by extending the time axis as a dimension of space
but still does not focus on their temporal changes or the system dynamics.
While RNNs have enabled dynamic processing, their learning methods still remain stuck in static approaches.
%Most conventional NN learning, which started with learning a static mapping from inputs to outputs
%with no time axis, focuses on the point and aims to move it in an appropriate direction
%based on an error function or some other static function of the outputs.
%The gradient method helps to determine the update of all the weights and biases.
%RNNs have enabled dynamic processing along the time axis, and BPTT has enabled learning of it.
%However, the aim is still to reduce some error function which does not take into account its temporal changes,
%and to move the point in the output space at the current step

However, since a state changes continuously over time, a point forms a line,
and changes in the neighborhood around that line form flows in the state space.
The point representing the current state is merely a cross-section for convenience
of what is essentially a continuous line and generally cannot be separated from the system's dynamics.
%unlike spatial information, time proceeds constantly in one direction.
In control systems, for instance,
%a controller cannot achieve good control by only using the current error from the target trajectory.
system stability is discussed based on convergence or divergence around the equilibrium point,
and for better control, the controller often uses the time derivative of the error from the target trajectory.

Dynamic Learning transforms neural network learning from ``learning of points'' to ``learning of flow,'' in other words,
from ``static'' to ``dynamic.''
Unlike control problems, where a target trajectory is specified, and stability is absolutely required,
%in RL, an agent does not have target trajectories.
an agent's behavior should be exploratory initially and then become less exploratory over time,
and the system dynamics must not fully converge.
Thus, as a concrete method for dynamic learning,
the author arrived at the idea of controlling the convergence or divergence of the flow directly
around the current transition, rather than moving points.
He also arrived at the idea of controlling global dynamics through controlling dynamics locally in both space and time.
The author believes he could demonstrate that these ideas work appropriately, at least in RL.
%The agent explores based on the chaotic system dynamics,
%and the chaoticity is controlled according to being better or worse of the current state transition.
%Many partifal differential equations in which the time axis is not symmetrical to any spatial axis
%are useful in expressing the relationship between space and time.
%Thus, time is not merely one of the axes in space, but should be taken as a special one.
%Then Dynamic Learning tries to control the convergence or divergence of the flow according to the gradient method.
%The difference appears in Fig.~\ref{fig:ConvRL} and Fig.~\ref{fig:DYN_RL}.
%In Fig.~\ref{fig:ConvRL}, the aim is to move the output vector at the moment, but in Fig.~\ref{fig:DYN_RL},
%the aim is to control the convergence or divergence of the neighborhood flow in the network state space over time.
%as weights and biases are updated to control the convergence or divergence of flows,
%the states and outputs themselves also change.
%Even when the neighborhood along a state transition becomes convergent through learning with a positive TD-error,
%the original state transition itself may change. 
%In a network with a large number of neurons, it is known that any ideal function exists in the neighborhood
%of any sufficiently wide random network, this new learning works better
%though the author cannot examine due to his poor computational resources.
%The author thinks that it can easily be extended to continuous time.
%It can also be extended to supervised learning by subtracting the temporal average of the error function from itself.

This shift in neural network learning is fundamental.
% and has the potential to be applied more widely.
However, using ideal divergence/convergence as supervised signals for learning dynamics does not seem very practical.
Though it may be possible to compute and use ideal dynamics derived from ideal outputs, which are given as training signals,
it remains unclear how much advantage this approach would have over direct learning with conventional static-type supervised learning.
It may be mainly used in RL at last.
However, in large-scale NNs, thanks to the magic of high-dimensional geometry,
appropriate parameters (weights and biases) for a given cost function do not lie so far from their initial values \citep{Amari}.
The author speculates that in large-scale RNNs, ``learning of flow'' along the time axis may become more significant and effective
than ``learning of point'' on a time cross-section,
though the scale of the RNN used in this study was not large enough to experience these benefits.

As noted in the Introduction, Dynamic RL brings about a big transformation to exploration.
Exploration goes from stochastic to deterministic and no longer requires a random number generator.
In Dynamic RL, exploration is not performed at the level of actions or final outputs
but at the level of the entire process within the agent's RNN.
This exploration is expected to be considerably more flexible because it is learnable and also has many DOFs
rather than relying on stochastic selection with a scalar parameter called temperature.
The excellent adaptability to a new environment, as seen in the simulation results,
may be considered an advantage of such high-performance exploration.
Furthermore, the author finally positioned exploration as a preliminary step to grow into `thinking.'

\begin{figure}[t]
\centerline{\includegraphics[scale=0.26]{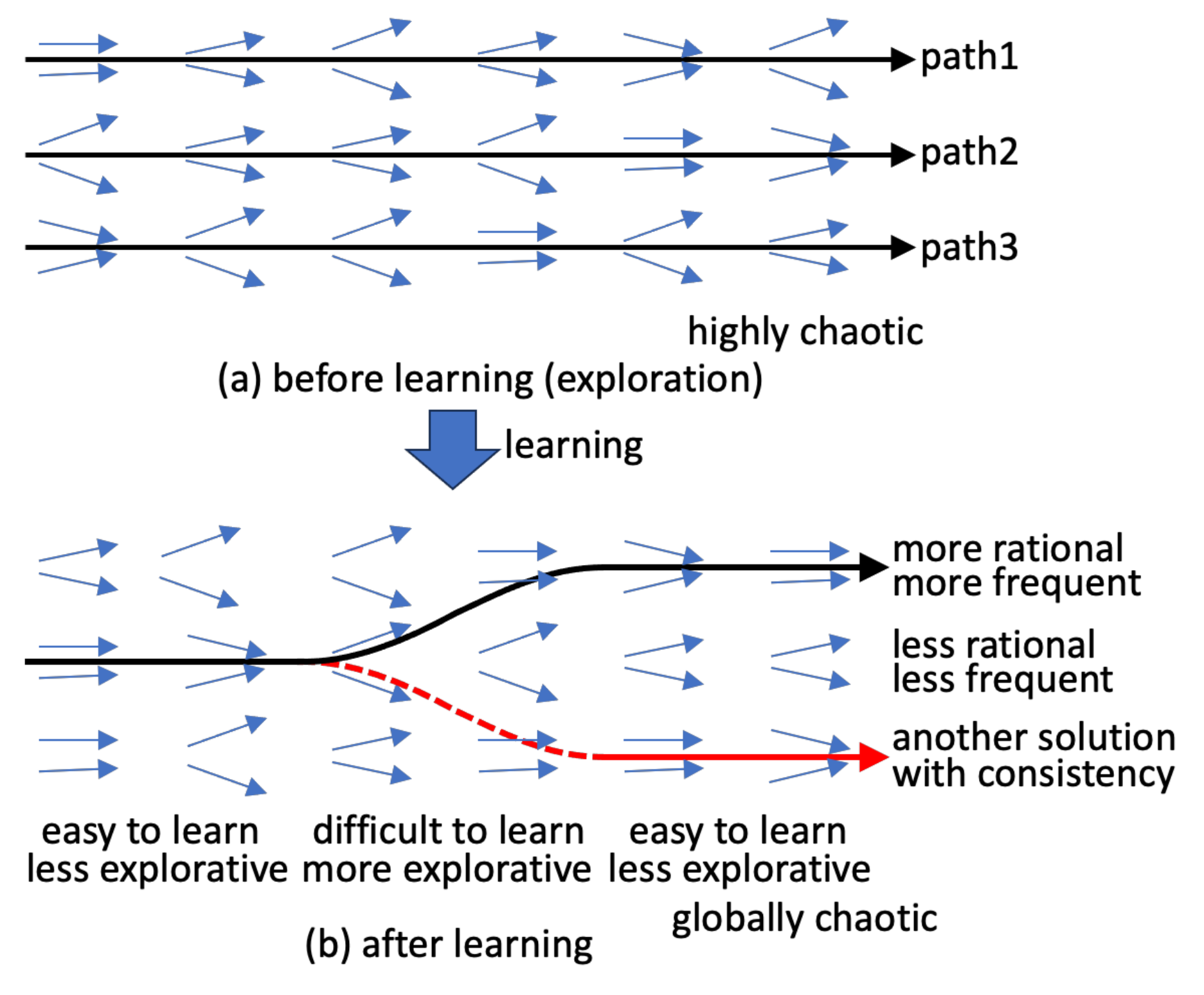}}
\caption{A conceptual diagram explaining how Dynamic RL works in simplified 2-D state space at the expense of exactness.
The reproducibility of better state transitions increases, and
novelty with consistency is expected to emerge as the red line after learning.}
\label{fig:LearningDynamics}
\end{figure}
Here, let us consider how Dynamic RL with such exploration works in a simplified two-dimensional state space as an example,
using a conceptual diagram, as shown in Fig.~\ref{fig:LearningDynamics}.
Before learning, dynamics are set to be highly chaotic, and state transitions are overall explorative.
Dynamics are learned around the actual paths depending on the TD-error at each state.
Assuming that there are regions in which learning is easy and a region in which learning is not easy in the state space.
In the former cases, by making the flows around rational state transitions with a positive TD-error converge,
the flows gather around such rational state transitions.
In the latter cases, the flows are still more explorative.
If it occurs in a high-dimensional space, pseudo-attractors would emerge.
Thus, as shown in Fig.~\ref{fig:LearningDynamics}(b),
a rational state transition, indicated by a thick black line with an arrow at the end, is achieved.
Furthermore, due to the explorative transitions in the region where learning is difficult,
sometimes another solution emerges.
However, the author expects the transitions not to become incoherent
because the state transitions are rational in the regions where certainty in learning is high.

%Dynamic RL brings a qualitative transformation to the exploration as well.
%Conventional exploration is performed
%as a stochastic selection after all the processes for the action or motion generation at each time step.
%The exploration here is not a stochastic process, but a deterministic process utilizing chaotic dynamics of the system.
%It is performed throughout the process of the agent.
%It is not static, focusing only on a specific point in time, but dynamic along the time axis.
%In the Introduction, the author touched that the high DOF exploration in Dynamic RL
%can balance the persistence and flexibility and promote to step outside of common sense.
%The author did not check the big merit here, but the excellent adaptation speed in Dynamic RL
%may be a fruit of such exploration.

\begin{figure}[t]
\centerline{\includegraphics[scale=0.36]{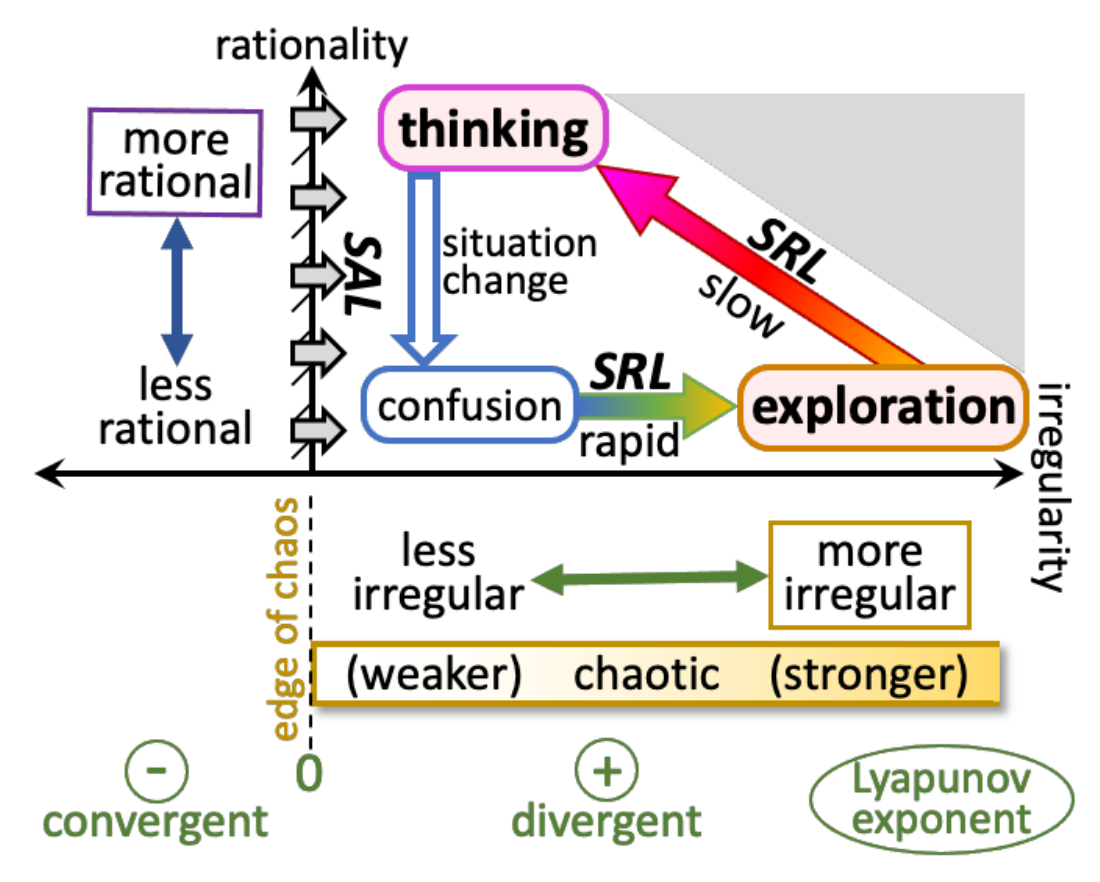}}
\caption{`Exploration' and `thinking' loop and its relationship to Dynamic RL, the author advocates.
This is an updated version of Fig.~\ref{fig:Thinking} and was obtained through the analysis of the simulation results.
SAL maintains chaotic dynamics of the system,
thereby preventing a decline in learning activities.
Under this condition, SRL changes the agent's processing from ``more irregular'' to ``more rational'' gradually;
in other words, it moves the processing from `exploration' to `thinking'.
When situations are changed, the state transitions become less rational for the agent,
and negative TD-errors appear more frequently or with larger magnitudes.
(labeled this as ``confusion'')
Then SRL makes the state transitions ``more irregular'' rapidly.
}
\label{fig:ThinkingExplLoop}
\end{figure}
Next, the author summarizes how he thinks about `exploration', `thinking,' and their relationship to Dynamic RL,
referring to Fig.~\ref{fig:ThinkingExplLoop}.
As mentioned, since both `exploration' and `thinking' need autonomous state transitions
without being stuck in a convergent state,
the system dynamics, including both the agent and its environment, should be chaotic.
It ensures the agent's activity in both behaviors and learning.
SAL plays this role (right-pointing gray block arrows in Fig.~\ref{fig:ThinkingExplLoop}).
The agent explores at first with strong chaotic system dynamics.
SRL enhances the reproducibility of the autonomous state transitions that are evaluated as better by the critic,
and thus, the transitions and also the agent behaviors become more rational (diagonal block arrow).
In this way, exploration grows into thinking, which is autonomous and more rational state transitions
with weak chaotic system dynamics.
This learning is slow because it takes time for the agent to explore rational state transitions,
in other words, to acquire the knowledge needed for getting rewards and avoiding punishments.

When the situation changes, such as the environmental change in the learning tasks in this paper,
the frequency or magnitude of worse evaluations, i.e., negative TD-errors, increases.
This means that the state transitions become less rational  (downward pointing block arrow).
Then, SRL resumes exploration autonomously by making the system dynamics more chaotic
(right-pointing large block arrow).
This learning progresses rapidly because it only requires increasing the sensitivity of each neuron.
The relationship between exploration and exploitation is quite similar to that between exploration and thinking.
Exploitation and thinking are similar in that both are rational.
Therefore, this mechanism can also explain how Dynamic RL resolves the ``exploration and exploitation'' dilemma
and balances them depending on the situation.

The brain, a massively parallel system, is too interconnected and complex for us to understand,
and we cannot easily divide its processing into modules.
For many years, the author has believed that our functions, such as recognition and prediction, are
merely convenient labels to understand the brain's complicated parallel processing by dividing it into parts \citep{Intech}.
For instance, processing performed close to sensors is commonly called ``recognition.''
We recognize an image while we think, ``Is this inside a house?''
``Is it daytime or nighttime?'' or ``What kind of people live there?'' drawing on our knowledge.
These thoughts occur during image recognition, 
making it difficult to classify them clearly as either thinking or recognition.
%and cannot be separated from recognition.
%In this way, recognition and thinking would be deeply connected and are not what can be divided clearly.

For ease of understanding, this paper has discussed exploration and thinking
as if they exist alongside other functions.
However, considering the actual human processing as described,
these should not be labels for any part of the overall process.
Instead, they may represent a tendency of our processing,
including recognition and motion generation, to be either more irregular or more rational.
In the presented simulations, sensor signals were used as input to the actor network without any preprocessing.
Hence, all processes from sensors to motors are performed in the network,
even though it is referred to as an ``actor.''
Learning was applied to the entire network, not just a part of it,
and exploration was generated throughout the system, including the external world.
This is consistent with the findings that chaotic attractors for existing odorants and chaos for new ones were observed
in the olfactory bulb \citep{Skarda, Freeman} even though it does not seem to be directly related to motion generation.
The relationship between Dynamic RL and the ``default mode network'' \citep{DMN} would also be interesting
in terms of thinking.

Supervised learning has an excellent ability to generate reasonable answers
through learning a vast amount of examples but originally does not have the ability to discover something new by itself.
RL receives no direct training signal; instead, it has the ability to discover new things through exploration.
Exploration in Dynamic RL is performed flexibly throughout the agent's RNN with massive DOFs.
Such exploration grows into `thinking' through learning in harmony with novelty and rationality, maintaining the massive DOFs.
Accordingly, the author believes that thinking that emerges in this way would make it possible to step outside of common sense
without becoming incoherent, which is difficult for current Transformer-based LLMs to achieve,
as introduced in the Introduction.

However, thinking seems to be a conscious process that is more sequential than parallel.
This may simply suggest that we can identify only the conscious part of the processes, which are just the tip of the iceberg.
Visible language-level processing is exactly what current LLMs excel at, as shown in the Introduction.
Given the LLMs' outstanding abilities, the author believes that integrating Dynamic RL with an LLM is essential to realize `thinking.'
\subsection{Risks and Concerns}\label{subsec:Risk}
Lastly, the author expresses concerns about the risks associated with this research.
%As mentioned, the author thinks that the frame of this research has the possibility to open the way to the emergence of thinking
%including inspiration or discovery.
%If artificial creatures begin to think autonomously and discover ideas beyond human imagination, they could pose a grave risk to humanity.
%
%However, if the author's concern were to come to pass, 
%
%AI weapons such as LAWS (Lethal Autonomous Weapons System) are now in actual use on the battlefield.
%https://docs-library.unoda.org/Convention_on_Certain_Conventional_Weapons_-Group_of_Governmental_Experts_on_Lethal_Autonomous_Weapons_Systems_(2023)/CCW_GGE1_2023_CRP.1_0.pdf
The risks of AI have been frequently pointed out \citep{Signatories, Doya, AI_Risk}.
%, especially AGI (Artificial General Intelligence), 
The general consensus on science and technology that we should benefit from them while avoiding their risks
has been widely accepted also in the AI community \citep{Signatories, Doya}.
%it is widely accepted in the AI community that
%``it is important to research how to reap its benefits while avoiding potential pitfalls'' 
%However, with the advent of generative AI, many ones feel the risk more real, and the word ``catastrophic risk" has been often seen. 
%Among them, some mention the responsibility of researchers,
%but the author is not aware of anyone other than YUdkowsky who has mentioned the stop of the research itself.
%Generally, researchers are held accountable for their own research,
%and people are encouraged to discuss establishing regulations on its use.
%but generally, the greatest responsibility lies with the users, not the researchers.
However, if AI entities begin to think autonomously and discover ideas beyond human comprehension,
%they could pose a grave risk to humanity.   % life-or-death
they could threaten humanity.
%the potential dangers of autonomous thinking may exceed our expectations.
%Currently,
%The author has no solution to ensure that we can control and avoid such risks.
In this context, it is groundbreaking that the prioritization of risk aversion over profit-taking is clearly stated in \citep{AI_Risk1} as follows:
``The potential benefits of AI could justify the risks if the risks were negligible.
However, the chance of existential risk from AI is too high for it to be prudent to rapidly develop AI.
Since extinction is forever, a far more cautious approach is required.''
As mentioned, the author believes that Dynamic RL's potential to discover new things would be significantly higher
than the other AI technologies and should be especially considered. 

%Of course, the author understands that the level of this research is not at all up to the level of concern at this moment and that
Of course, given the current level of this research, these concerns must seem presumptuous or alarmist.
%the author acknowledges.
%Technological advancement, such as the performance of semiconductors, often follows an exponential curve \citep{Kurzweil}.
However, once AI entities reach a level where they can design more intelligent versions of themselves
without any human assistance, their growth will surpass exponential rates,
unlike human-developed technologies such as semiconductors \citep{Kurzweil}.
It will accelerate far beyond our predictions.
Above all, they will get out of control and evolve rapidly even if we stop the research,
leaving us with no means to avoid or halt them.
%However, once AI can think by itself with its own will without human intervention, how may humans stop it?
%it will not only make its own decisions but also design itself.

The author begs people to seriously imagine the horror of autonomous, superintelligent drones flying around
and humanoids walking around and targeting humans as a potential reality. 
He has no idea how we can absolutely ensure
that superintelligent drones or humanoids will never muzzle humanity.
%The author cannot understand the rationale of a foolproof scenario
%in which superintelligent drones would never turn against humanity.
%Once AI reaches a level where it can design higher levels of intelligence without any external help,
%its growth will accelerate exponentially, far beyond our imagination when we realize that.
%Then, humans may have no means to stop it, leaving them with no option but to fight.
%Then, humans may have no means to stop it, except for fighting and extinguishing superintelligence.
%we will no longer be able to stop it.
%The author guesses this would be what A.~Turing wanted to say in 1950 \citep{Doya}.
% presumptuous 厚かましい，僭越な，おこがましい
% seems conceited 自惚れた
% sheer arrogance 傲慢
Given the gravity of the matter and the fact that redoing it is not an option, 
easy wishful thinking without a thorough understanding should definitely be avoided \citep{Yudkowsky}.
%The author does not have a clear vision of its future development,
%With advancements in science and technology allowing us to live satisfactorily,
Science and technology have made remarkable progress.
Now, they have reached a point where we can lead satisfactory lives
although challenges remain in other aspects, such as wars and economic disparities.
% by utilizing them, thanks to our predecessors for their accumulated efforts,
%What kind of benefit on earth is worth putting humanity's safety at a fatal risk?
What kind of benefit on earth justifies putting humanity's safety at such a fatal risk?
We must seriously consider this.
% In today's society, which is already very convenient,

History shows that 
%once the potential of technology is unveiled, it spreads rapidly. Moreover, 
once nations begin developing lethal weapons using a new powerful technology,
it becomes extremely difficult to eliminate them in power games, as in the case of nuclear weapons.
%The imagination and wisdom of humanity are being tested.
However, the author would like to strongly warn that
%, unlike the case of nuclear weapons,
if one nation attempts to use this technique to conquer another, 
it is likely that the world, including the nation itself, will eventually be defeated by the technique.

The researchers developing technologies can usually be the first to recognize their potential risks
by using their imagination about what will happen.
Given the severity of the potential consequences and the importance of early action,
they should not throw the decision solely to society but must take responsible measures themselves first.
The author may have neither the right nor the power to halt the research.
Submitting this research, despite being aware of its potentially fatal risks, represents a major contradiction.
%It is a major contradiction to the submission of this research, which he believes can pose fatal risks.
Some may also argue that such discussions should be excluded from a technical paper.
However, even without this submission, someone else will eventually conduct this research.
The author believes that we have entered an era
where researchers must take a certain amount of responsibility for their research as soon as they publish a paper on it.
Therefore, in order to raise the issue and spark discussions that harness the collective wisdom of humankind,
the author takes the bold step of making this submission and hereby presents the following statement
to initiate discussions in order to fulfill his responsibilities as a researcher.

%\begin{itembox}[|]{\bf Statement}
%\noindent{\bf Statement for Discussion}\\
%Dynamic RL proposed here has the potential to endow AI including AGI (Artificial General Intelligence)
%with the ability to think and discover new things, which could pose a grave threat to humanity.
%Now, while this RL research is still in its baby steps,
%we should halt its further progress to protect humanity from the risk.
%%for the sake of humanity.
%The author sincerely and earnestly urges researchers to refrain from advancing this work
%and developers to avoid developing a library for Dynamic RL in frameworks,
%% not to proceed with
%at least until a consensus is reached among humans on this matter.
%Let us pause, imagine, and contemplate, free from preconception before it is too late.
%\end{itembox}

%\noindent
%Dear reviewers,\\
%If you think the above description of what the author thinks about the risk of this research is inappropriate for a technical paper, could you point out concretely why and which part is inappropriate, please?

\section*{Acknowledgment}
First, I would like to express my gratitude to Prof.~Kazuyuki Aihara.
His suggestion in 1995 motivated me to connect reinforcement learning and chaos.
%I would also like to thank Prof. Kanakubo from the Shizuoka Inst. of Sci. \& Tech. for their impressive demonstration of chaotic itinerancy.
I would like to thank Prof.~Hiromichi Suetani for useful suggestions about chaotic dynamics from an expert.
I also thank the members of our former laboratory for their discussions and also sharing related simulation results.
In particular, Dr.~Toshitaka Matsuki discussed mainly the relationship between chaoticity and learning ability
and has inspired me a lot from various aspects.
Mr.~Yuki Tokumaru also gave me a big push by simulations and discussion in the early phase of this research.
This work was supported by JSPS KAKENHI Grant Number JP19300070, JP23500245, JP15K00360, JP20K11993
and also Kayamori Foundation of Informational Science Advancement K31-KEN-XXIV-539.
Especially the last one supported the author's independent research after his early retirement.

%\section*{Declaration of Generative AI and AI-Assisted Technologies in the Writing Process}
%During the preparation of this work, the author used ChatGPT and Gemini, in order to assist with improving his English writing.
%After using this tool, the author reviewed and edited the content as needed and takes full responsibility for the content of the publication.

%\section*{References}
\bibliographystyle{abbrvnat}
%\bibliographystyle{agsm}
%\bibliographystyle{apa}
%\bibliographystyle{apalike}
%\bibliographystyle{unsrt}
%\bibliographystyle{abbrv}
%\bibliographystyle{plain}
%\bibliography{mybibfile}
\bibliography{Reference}

\par
\par
\par

\appendix
\section{Detailed Task Description for Sequential Navigation Task}\label{App:Task}
\noindent
{\bf Initial agent location}\\
Once per four episodes: randomly chosen on the four sides of the $18\times 18$ square
whose center is on the origin to promote the learning in the peripheral area.\\
Other episodes: randomly chosen in the whole field.\\
{\bf Visual sensor}\\
Number of cells: $11 \times 11=121$. 
Cell size: $1\times1$.
The sensor is fixed with its center at the origin, and the cells are arranged in a square without overlap.
Appearance of agent: $1\times1$ square.\\
Cell output: the area occupied by the agent as a valued from 0.0 to 1.0.\\
{\bf Agent's one-step move}\\
Each actor output $\rightarrow$ multiplied by 1.25 $\rightarrow$ clipped between -1.0 and 1.0.
Furthermore, keeping the direction of the move vector, the size of the vector is normalized
so as that the maximum size for the direction is 1.0.
For example, if the actor outputs are (0.0, 0.9), the agent will move by (0.0, 1.0),
and if they are (0.3, 0.4), it will move by $(0.3, 0.4)$.
As a result, the one-step movable area becomes a circle with a radius of 1.0.\\
{\bf Failure criteria}\\
The average number of steps to the goal was calculated for every 500 episodes.
The simulation run was terminated when one of the following conditions was satisfied.
(1) ($episode\_No. \geq 5000\ \&\ average >190$) for 5 consecutive times
or (2) ($episode\_No. \geq 10000\ \&\ average > 150$) for 5 consecutive times
or\\.                               (3) ($episode\_No. \geq 10000\ \&\ average > 100$) for 10 consecutive times

\section{Initial Weights and Learning Rate}\label{App:Table}
The parameters used in this paper are detailed further in Tables \ref{Table:WeightsAndBiases}.
\begin{table}[h]
  \caption{Parameters for Critic and Actor networks.
  When Dynamic RL was used in the actor network, only the learning rate for SRL is written.
  The learning rate for SAL was always set as $\eta_{SAL} = \eta_{SRL}*0.1$.
  in, hid, hid1, hid2, out: input, hidden, lower hidden, upper hidden, output layer respectively.
  In the column of ``Initial value'', N: Normal-distributed random number (R: Spectral radius),
  Number only: Uniform random number (Its absolute value is less than the value).}
  \label{Table:WeightsAndBiases}
  \centering \small
  \vspace{5mm}
  Critic network\\
  \vspace{3mm}
   \begin{tabular}{cccccc}
  \hline
                       & layer $\rightarrow$ layer & \multicolumn{2}{c}{Initial value} & \multicolumn{2}{c}{Learning rate}\\
                       &  or layer                           & Switch & Crank & Switch & Crank\\                       
  \hline                     
  Weight & in $\rightarrow$ hid & 0.2 & 0.5 & 0.2 & 0.3\\
                       & hid $\rightarrow$ hid & \multicolumn{2}{c}{N R1.3} & \multicolumn{2}{c}{0.002}\\
                       & hid $\rightarrow$ out & \multicolumn{2}{c}{0.0} & \multicolumn{2}{c}{0.02}\\
  \hline
  Bias& hid, out & \multicolumn{2}{c}{0.0} & \multicolumn{2}{c}{0.02}\\
  \hline
  \end{tabular}
 
  \vspace{5mm}
  Actor network (Dynamic RL)\\
  \vspace{3mm}
  \centering \small
  \begin{tabular}{cccccc}
  \hline
                      &   layer $\rightarrow$ layer  & \multicolumn{2}{c}{Initial value} & \multicolumn{2}{c}{Learning rate $\eta_{SRL}$}\\
                       &  or layer                             & Switch & Crank & Switch & Crank\\                       
  \hline                     
  Weight & in $\rightarrow$ hid & 0.2 & 0.5 & 0.01 & 0.005\\
                       & hid1,2 $\rightarrow$ hid2,1 & \multicolumn{2}{c}{0.1} & 0.005 & 0.002\\
                       & hid2 $\rightarrow$ hid2 & \multicolumn{2}{c}{N R3.0} & 0.005 & 0.002\\
                       & hid1 $\rightarrow$ out & \multicolumn{2}{c}{0.1} & 0.01 & 0.005\\
  \hline
  Bias& hid1, hid2 & \multicolumn{2}{c}{0.0} & 0.002 & 0.0002\\
         & out  & \multicolumn{2}{c}{0.0} & 0.002 & 0.0005\\
  \hline
  \end{tabular}
  
  \vspace{5mm}
  Actor network (conventional RL)\\
  \vspace{3mm}
  \centering \small
  \begin{tabular}{cccccccc}
  \hline
                      &   layer $\rightarrow$ layer & \multicolumn{2}{c}{Initial value} & \multicolumn{4}{c}{Learning rate}\\
                       &  or layer                           & Switch & Crank & \multicolumn{2}{c}{Switch} & \multicolumn{2}{c}{Crank}\\                       
                       &                                         &         &        &  BPTT& SAL  & BPTT& SAL\\                       
  \hline                     
  Weight & in $\rightarrow$ hid & 0.2 & 0.5 & 0.1 & 0.01 & 0.2 & 0.01\\
                       & hid1,2 $\rightarrow$ hid2,1 & \multicolumn{2}{c}{0.1} & 0.01 & 0.01 & 0.01 & 0.005\\
                       & hid2 $\rightarrow$ hid2 & \multicolumn{2}{c}{N R1.3} & 0.002 & 0.01 & 0.002 & 0.005\\
                       & hid1 $\rightarrow$ out & \multicolumn{2}{c}{0.0} & 0.01 & - & 0.01 & -\\
  \hline
  Bias& hid1 & \multicolumn{2}{c}{0.0} & 0.002 & 0.01 & 0.002 & 0.002\\
         & hid2 & \multicolumn{2}{c}{0.0} & 0.001 & 0.01 & 0.001 & 0.002\\
         & out  & \multicolumn{2}{c}{0.0} & 0.001 & - & 0.001 & -\\
  \hline
  \end{tabular}
\end{table}
%\begin{table}[t]
%  \caption{Other parameters}
%  \label{Table:OtherParameters}
%  \centering \small
%  \begin{tabular}{ccc}
%  \hline
%  & Switch Task & Crank Task\\
%  \hline
%  Discount factor $\gamma$ for critic in Eqs.~(\ref{Eq:TDerr}),(\ref{Eq:C_train})& \multicolumn{2}{c}{0.98}\\
%  Rate of regulation $\eta_{reg}$ in Eq.~(\ref{Eq:Regularize}) & $1E^{-6}$ & $1E^{-7}$\\
%  Moving average for sensitivities $\alpha$ in Eq.~(\ref{Eq:Ave_sen}) &  \multicolumn{2}{c}{0.001}\\
%  Moving average for Raising critic $\beta$ in Eq.~(\ref{Eq:Ave_C}) &  \multicolumn{2}{c}{0.0001}\\
%  Threshold of Raising critic $C_{th}$ in Eq.~(\ref{Eq:Raise_Critic}) & \multicolumn{2}{c}{0.1}\\
%  Rate of raising critic $\eta_{raise}$ in Eq.~(\ref{Eq:Raise_Critic}) & 0.0005 & 0.0002\\
%  SAL Target $s_{th}$ in Fig.~\ref{fig:DynamicRL} & 1.3 & 1.6\\
%  %Learning rate for SAL & \multicolumn{2}{c}{0.1}\\
%  Truncated steps in BPTT & 20 & 10\\
%  \hline
%  \end{tabular}
%\end{table}

\section{Computation of Exploration Exponent}\label{App:Expl_factor}
Before each episode, an agent is replicated with the same actor and critic networks as the original.
The original agent was placed in one of the nine starting positions, which are the same as those shown in Fig.~\ref{fig:Task1_Loci}(b) or (d).
The replicated agent was placed randomly at a position $10^{-6}$ units away from the original one.
The two agents move according to their actor outputs without learning until one of them finishes its episode.
Subsequently, the Euclidian distance between their states is observed, and the equation below is calculated.
Here, the agent's state is represented by the direct product space (222 dimensions)
of the input signal space (122 dimensions) and the upper hidden layer state space (100 dimensions).
%the inputs of the first hidden layer with 222 dimensions are used as the agent's state.
%The state of the system is defined as a point in the input signal space of the lower hidden layer with 222 dimensions.
%That is the sum space of the sensor signal space (122 dimensions)
%and the space of the neuron outputs in the upper hidden layer (100 dimensions).
%The Euclidian distance $\delta$ between the corresponding states between the two episodes is observed.
This was repeated for the nine start positions, and the exploration exponent was averaged across the nine sets.

The exploration exponent is defined as 
\begin{equation}
\lambda = \frac{1}{T_{end}-T_{start}}\sum_{t=T_{start}+1}^{T_{end}} ln \frac{\delta_{t}}{\delta_{t-1}}
=\frac{1}{T_{end}-T_{start}}ln\frac{\delta_{T_{end}}}{\delta_{T_{start}}}
\end{equation}
where $\delta_t$:  the distance between the two states at the $t$-th step, $T_{start}$: the start step
after the preparation steps, $T_{end}$: the final step in the episode.
When the distance $\delta$ is greater than $10^2$ or less than $10^{-10}$,
$T_{end}$ was set to the previous step number.
When the agent went to a corner of the field, the locations often became exactly the same
between the two episodes,
but the computation continued as long as the distance remained in the range of $10^2$ to $10^{-10}$.

\section{Equation of Motion of Slider-Crank Mechanism}\label{App:Eq_Crank}
As the equation of motion of the slider-crank mechanism, as shown in Fig. \ref{fig:Crank},
the author used the following equation
\begin{equation}
J\ddot{\theta} = RF_{\theta} - D\dot{\theta}\\
\end{equation}
where
\begin{equation}
F_{\theta} = f \frac{sin(\phi+\theta)}{cos\phi}\\
\end{equation}
\begin{equation}
cos\phi = \frac{L^2+x^2-R^2}{2Lx}\\
\end{equation}
\begin{equation}
x = Rcos\theta + \sqrt{L^2 - R^2sin^2\theta}\\
\end{equation}

%\allowdisplaybreaks

\begin{align}
J&: \text{the moment of inertia of the rotor} (=10.0)\nonumber \\
\theta&: \text{the rotational angle of the rotor}\nonumber \\
R&: \text{the distance from the center of the rotor to the crank} (=1.0)\nonumber \\
D&: \text{the damping coefficient for the rotation of the rotor} (=0.1)\nonumber \\
F_{\theta}&: \text{the force applied to the rotor in the direction of its rotation}\nonumber \\
f &: \text{the force applied laterally to the end of the rod, determined by the actor's outputs}\nonumber \\
\phi&: \text{the angle of the rod}\nonumber \\
L&: \text{the length of the rod} (=3.0)\nonumber \\
x&: \text{the distance from the center of the rotor to the rod end}\nonumber
\end{align}

\end{document}